\documentclass[lettersize,journal]{IEEEtran}
\usepackage{amssymb}
\usepackage{amsmath,amsfonts}
\usepackage{algorithmic}
\usepackage{algorithm}
\usepackage{array}
\usepackage[caption=false,font=normalsize,labelfont=sf,textfont=sf]{subfig}
\usepackage{textcomp}
\usepackage{stfloats}
\usepackage{url}
\usepackage{verbatim}
\usepackage{graphicx}
\usepackage{cite}
\usepackage{colortbl}
\usepackage{capt-of}
\usepackage{xcolor}

\begin{document}
\title{Disentangled Graph Representation Based on Substructure-Aware Graph Optimal Matching Kernel Convolutional Networks}
\author{Mao~Wang,~\IEEEmembership{}
	Tao~Wu,~\IEEEmembership{}
	Xingping~Xian,~\IEEEmembership{}
	Shaojie~Qiao,~\IEEEmembership{}
	Weina~Niu,~\IEEEmembership{}
	Canyixing~Cui~\IEEEmembership{}}
\maketitle

\begin{abstract}
Graphs offer an effective characterization of relational data in various domains, giving rise to graph representation learning that discover the underlying information for prediction. Graph Neural Networks (GNNs) have emerged as the state-of-the-art method, providing an end-to-end learning strategy for different tasks. Recently, disentangled graph representation learning has been proposed to enhance representation quality by decoupling the independent interpretable factors within graph data. However, existing methods still implicitly and coarsely characterize graph structures, limiting the analysis of structural patterns in graphs. Accurately capturing structural information of graphs has become the key to disentangled graph representation learning. This paper views the graph as a set of node-centric subgraphs, each serving as the structural factor to characterize specific information at various positions. In this way, graph-oriented predictions are transformed into the problem of pattern recognition for structural factors. Inspired by traditional Convolutional Neural Networks (CNNs), this paper proposes the Graph Optimal Matching Kernel Convolutional Network (GOMKCN) to learn disentangled representations for individual factors. A novel Graph Optimal Matching Kernel (GOMK) is designed as a convolutional operator to calculate the similarity between each subgraph and several learnable graph filters. Mathematically, GOMK maps each subgraph or graph filter into a point in Hilbert space, representing the entire graph by a set of points. The disentangled representation of a structural factor is equal to the projections of its subgraph onto individual graph filters. Driven by the gradient of training loss, these graph filters will capture task-relevant structural patterns. Additionally, GOMK considers local correspondences between graphs in the similarity measurement, resolving the contradiction between differentiability and accuracy in existing graph kernels. Experiments demonstrate that the model enables accurate graph pattern mining and achieves excellent accuracy and interpretability in prediction. This paper offers a new theoretical framework for disentangled graph representation learning.
\end{abstract}


\section{Introduction}
\IEEEPARstart{G}{raph} is universally applied to represent structured data across various domains\cite{essam1970some}, including biochemical and social networks. This has led to the development of graph representation learning, which aims to extract the underlying information from the graph and project it into vectors for prediction\cite{hamilton2020graph},\cite{zhang2018network}. Early works, such as Deepwalk\cite{perozzi2014deepwalk}, Line\cite{tang2015line}, and node2vec\cite{grover2016node2vec}, generate the representation for individual nodes, with the distribution keeping the neighbor relationships. However, these methods ignore the node features and fail to capture high-order structural information in the graph. With the development of deep learning techniques, Graph Neural Networks (GNNs) have been proposed for end-to-end graph representation learning. Generating the node representations with a message-passing mechanism\cite{gilmer2017neural}, they are widely used in fraud detection\cite{pourhabibi2020fraud}, drug discovery\cite{ma2018drug}, and traffic flow forecasting\cite{li2021spatial}. Although GNNs, e.g., GCN\cite{kipf2016semi}, GAT\cite{velivckovic2017graph}, GraphSAGE\cite{hamilton2017inductive}, have become the state-of-the-art method in predictions, it is discovered that the representations provide a coarse and implicit characterization of the graph\cite{yang2018node}.

\begin{figure}[t]
	\centering
	\includegraphics[width=3.5in]{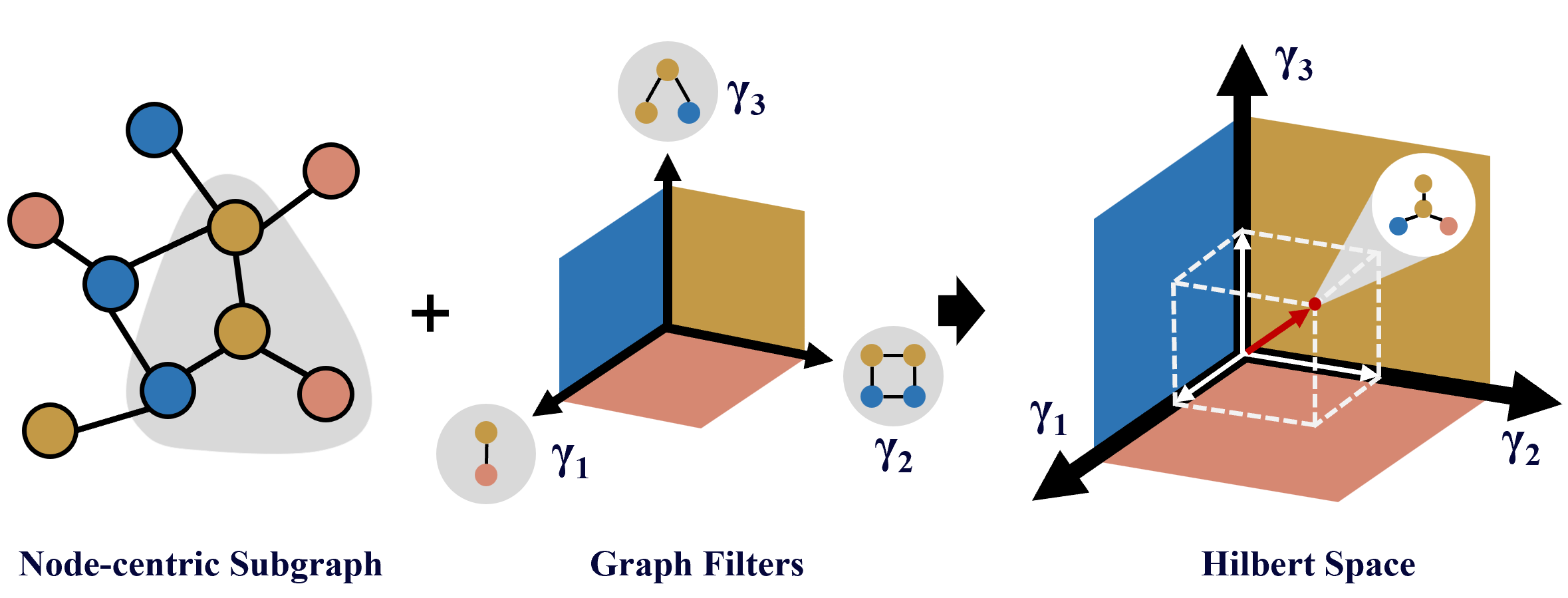}
	\caption{The idea of decoupled representation learning. A graph can be represented as a set of node-centric subgraphs. By mapping these subgraphs into a Hilbert space, representations of individual structural factors and the entire graph can be obtained.}
	\label{decouple}
\end{figure}

To improve the quality of representation, disentangled representation learning has been proposed to learn embeddings that denote mutually independent explanatory factors\cite{bengio2013representation}. Integrating this idea with the GNNs, research on disentangled graph representation learning has subsequently been initiated. DisenGCN\cite{ma2019disentangled} pioneeringly proposes a novel neighbor routing mechanism to identify latent factors behind the graph, thereby generating disentangled node representations. Based on this framework, IPGDN\cite{liu2020independence}, LGD-GCN\cite{guo2021lgd}, and FactorGCN\cite{yang2020factorizable} propose some improved methods and achieve promising results. Besides, a few works perform disentangled graph representation learning from a structural perspective. DisC\cite{fan2022debiasing} suggests that a graph is usually coupled with the causal subgraph and bias subgraph, contributing to genuine and spurious correlations in prediction, respectively. So, it strives to disentangle the latent variables by splitting the graph into causal and biased subgraphs. Recently, KerGNN\cite{feng2022kergnns} and RWK$^{+}$CN \cite{lee2024descriptive} propose to learn node representations by performing convolution on the graph. Yet, the model's performance is greatly limited by inaccurate graph pattern matching.

Although various methods of disentangled graph representation learning have been proposed, research in this field faces three challenges: \textbf{1. How to achieve disentangled representation learning based on structural patterns.} It is found that the vital characteristics and prominent patterns of a graph are reflected through local structures such as motifs\cite{milo2002network} and subgraphs\cite{ullmann1976algorithm},\cite{bouritsas2022improving}. However, existing disentangled graph representation learning primarily focuses on node features while neglecting the structural patterns, resulting in the inability to distinguish the structural patterns in the graph accurately. \textbf{2. How to adaptively discover the target structural pattern end-to-end.} Some works attempt to enumerate motifs and perform matching within the graph to extract structural representations\cite{yang2018node},\cite{chen2023motif}. Nevertheless, faced with different datasets and tasks, the critical patterns in the graph may differ. So, those enumerated motifs may fail to capture critical structural patterns without an end-to-end training strategy. \textbf{3. How to address the contradiction between differentiability and accuracy in graph pattern matching.} The structure-oriented representations heavily rely on recognizing structural patterns through graph pattern matching. To this end, various graph kernels have been designed to measure graph similarity. However, existing graph kernels can be classified into two categories: those based on $R$-convolution frameworks\cite{haussler1999convolution} and those based on the Label Refinement\cite{shervashidze2009fast},\cite{hido2009linear}. The former tend to introduce spurious similarities in their measurement\cite{kriege2012subgraph},\cite{kashima2003marginalized}, and the latter are non-differentiable\cite{shervashidze2011weisfeiler},\cite{kriege2016valid}. Thus, a graph kernel that balances differentiability and accuracy is essential for disentangled graph representation learning.

Here, we share the main ideas for solving the above challenges separately, as shown in Fig.\ref{decouple}. (1) We view a graph as a set of subgraphs centered around each node. These subgraphs capture the information from two aspects: the role of individual nodes in the coupling relationships and the specific details of the graph in various locations. It's worth noting that the mapping between the graph and the set can be bijective. Therefore, each subgraph can be regarded as a structural factor of the graph. (2) Almost any prediction task can be translated into the problem of recognizing and counting specific structural factors. Therefore, we introduce several graph filters to capture the critical structures for prediction. In particular, each graph filter's edge weights and node features are learnable parameters, and the disentangled node representations are generated by matching their subgraphs with these graph filters. As a result, inference based on these representations drives the graph filters to converge on critical structural patterns. (3) As for comparing two graphs, we draw inspiration from the $R$-convolution framework and decompose a graph into a collection of substructures. Differently, we generate an embedding of each substructure in a differentiable manner and attempt to establish an optimal matching between these substructures. This improves accuracy, differentiability, and interpretability in graph pattern matching.

This paper proposes a graph-kernel-based convolutional framework named GOMKCN for disentangled graph representation learning. Specifically, it generates the decoupled representation for each node by matching its subgraph with a series of trainable graph filters in the model. Herein, a graph kernel GOMK is proposed as the convolutional operator, calculating the similarity between the subgraphs and the graph filters. Mathematically, the representation of each node is equivalent to the projection of its subgraph onto the basis defined by the graph filters within the Hilbert space. For the comparison between a subgraph and a graph filter, the GOMK decomposes them into sets of subtrees rooted at individual nodes, respectively. Then, it utilizes a $t$-level Subtree Encoding ($t$-SE) to produce the embedding of the individual subtree. As a result, each subgraph or filter can be denoted by the collection of subtree embeddings. Subsequently, an Optimal Matching Kernel (OMK) is proposed to measure the similarity between two sets, which aligns the elements across sets and calculates the set similarity by solely focusing on the matched element pairs. Based on $t$-SE and OMK, the GOMK possesses differentiability, accuracy, and interpretability characteristics. This greatly facilitates the structural pattern mining in the graph and provides a solid foundation for constructing interpretable graph neural networks. A summary of our contributions is as follows:

\begin{itemize}
	\item We propose a theoretical framework named GOMKCN, which performs disentangled graph representation learning based on structural patterns. It equivalently decomposes a graph into a collection of node-centric subgraphs and obtains the disentangled representations by mapping them into Hilbert space.
	\item We propose the $t$-SE, a method of graph embedding that encodes a graph into a set of subtree embeddings. A mathematical analysis is conducted to ascertain its expressive power.
	\item We propose the OMK to measure the similarity of two sets with the consideration of optimal matching between their elements. Theoretical proof substantiates the efficacy of OMK to serve as a kernel function.
	\item Based on the $t$-SE and the OMK, we derived the GOMK to serve as the convolutional operator in GOMKCN. Experimental results demonstrate that it enables accurate graph pattern learning and provides excellent accuracy and interpretability in prediction.
\end{itemize}

The remaining content of the paper is organized as follows: Section \uppercase\expandafter{\romannumeral2} introduces related works on graph representation learning, disentangled graph representation learning, and graph kernels. Section \uppercase\expandafter{\romannumeral3} formally establishes the problem definition and presents necessary graph-theoretic preliminaries. Section \uppercase\expandafter{\romannumeral4} introduces the proposed GOMKCN framework, detailing its architectural components through three core operational phases: subgraph extraction, subgraph embedding, and subgraph matching. Section \uppercase\expandafter{\romannumeral5} conducts a rigorous theoretical analysis of GOMKCN's properties, with particular emphasis on the expressiveness of $t$-SE and the feature map of GOMK.  \uppercase\expandafter{\romannumeral6} describes the experimental methodology and presents the experimental results. Section \uppercase\expandafter{\romannumeral7} concludes with a synthesis of key contributions and suggests directions for future research.

\section{Related work}
\subsection{\textbf{Graph Represention Learning}}
Graph representation learning\cite{hamilton2020graph},\cite{zhang2018network} aims to extract implicit information from structured data and transform it into vectors, facilitating subsequent reasoning tasks. Initially, Deepwalk\cite{perozzi2014deepwalk}, Line\cite{tang2015line}, and Node2vec\cite{grover2016node2vec} propose preserving the neighborhood relationships within the distribution of representations. Inspired by word embedding technology, they generate random walk sequences on the graph and utilize the co-occurrence of nodes to produce representations. Subsequently, GNNs are proposed to learn representations, utilizing both the topological structure and the node features for collaborative prediction. They share a mechanism of message-passing\cite{gilmer2017neural} to generate node embeddings. Herein, GCN\cite{kipf2016semi} generates the representation for each node by iteratively aggregating the node features of its neighbors; GAT\cite{velivckovic2017graph} introduces an attention mechanism to perceive the different importance of neighbors in the aggregation stage; GraphSAGE\cite{hamilton2017inductive} performs neighbor sampling to adapt to large-scale graph representation learning; and GIN\cite{xu2018powerful} optimizes the aggregation function to enhance the expressive power of the model. However, these methods focus on node-level embedding, which is insufficient for structural analysis. Moreover, their leverage of network structures is implicit and coarse\cite{yang2018node}.

\subsection{\textbf{Disentangled Graph Represention Learning}}
Disentangled graph representation learning works on extracting explanatory factors from the structured data. 
Pioneeringly, DisenGCN\cite{ma2019disentangled} suggests that the graph is formed by the interaction of multiple latent factors. So, it proposes a novel neighborhood routing mechanism to identify the individual factors and learn disentangled node representations. Based on this framework, various methods have been proposed to improve the model's performance. IPGDN\cite{liu2020independence} introduces the Hilbert-Schmidt Independence Criterion to enhance the independence of node representations concerning different latent factors. LGD-GCN\cite{guo2021lgd} integrates local and global information of the graph to improve the intra-factor consistency and the inter-factor diversity of node representations. FactorGCN\cite{yang2020factorizable} decomposes the graph into several factorized graphs to disentangle multiple relations. IDGCL\cite{li2022disentangled} achieves self-supervised representation learning by comparing the original and augmented graph in the latent subspaces. NED-VAE\cite{guo2020interpretable} utilizes a variational autoencoder to extract representation for the latent factors in edge-related, node-related, and node-edge-joint. In addition, there is also work focusing on the disentanglement of substructures within the graph. DisC\cite{fan2022debiasing} tries to disentangle the causal and bias subgraph in the graph to improve the model's generalization capability. These methods focus on disentangling node features based on pairwise relationships, resulting in learning node-level representations that are insufficient to characterize the structural patterns.

\subsection{\textbf{Graph Kernels}}
Graph kernels are functions that quantify the similarity between graphs. Most existing graph kernels are designed using the $R$-convolution framework. This framework decomposes a graph into a specific set of elements, such as subtrees\cite{mahe2009graph}, subgraphs\cite{kriege2012subgraph}, random walks\cite{kashima2003marginalized}, \cite{zhang2018retgk}, shortest paths\cite{borgwardt2005shortest}. Then, a base kernel function is employed to compute the similarity between the elements, and the overall similarity between two graphs is the sum of the similarities for all pairs of elements across the two graphs. However, all graph kernels based on the $R$-convolution framework ignore the correspondence of components across the graphs, introducing similarities between mismatched components as noise. Among these graph kernels, the random walk kernel is particularly notable and widely used, which evaluates the similarity between two graphs by the number of identical random walks within a specified number of steps. This count can be efficiently computed using the direct product graph\cite{gartner2003graph}, constructed from the adjacency matrices of the two graphs. However, all the random walk kernels have a significant drawback: they allow the same node to be visited multiple times, leading to spurious similarities between graphs\cite{mahe2004extensions}.

There are also graph kernels beyond the $R$-convolution framework that demonstrate excellent performance. Weisfeiler–Lehman Subtree Kernel\cite{shervashidze2011weisfeiler} and Weisfeiler–Lehman Optimal Assignment Kernel\cite{kriege2016valid} measures the similarity between graphs by counting the number of nodes with the same labels in both graphs. These labels include both the original node labels and the generated labels through the mechanism of Label Refinement\cite{shervashidze2009fast},\cite{hido2009linear}. In particular, the new label of a node is generated by aggregating its own and the neighbors' labels and encoding them into a new label using a hashing algorithm. However, this mechanism is unsuitable for scenarios where nodes possess continuous numerical features. The lack of differentiability restricts its integration in GNNs.

Among all existing methods for graph representation learning with graph kernels, the works closest to this paper are KerGNN\cite{feng2022kergnns} and RWK$^{+}$CN \cite{lee2024descriptive}. They improve the random walk kernel to meet the differentiability requirement and then use it as the convolutional operation in the models. However, their measurement of graph similarity is still not accurate enough, while the GOMK we proposed tends to obtain a more precise similarity by aligning the two graphs.

\begin{table}[t]
	\caption{notations frequently used in this paper.\label{tab:notations}}
	\centering
	\begin{tabular}{cc}
		\hline
		Notations & Descriptions\\
		
		\hline
		$G$ & A Graph.\\
		$\mathbf{A}_{G}$ & Adjacency matrix of G.\\
		$\mathbf{F}_{G}$ & Feature matrix of G.\\
		$\mathcal{N}(u)$ & A set consisting of $u$'s neighbor nodes.\\
		$G_{u}^{k}$ & Subgraph centered on $u$ within $k$ hops.\\
		$\gamma$ & A set of graph patterns.\\
		$\kappa$ & A graph kernel.\\
		$\mathbf{z}_{u}$ & Disentangled representation of node $u$.\\
		
		\hline
		$\mathbf{f}_{v}$ & Original features of node $v$.\\
		$\mathbf{f}_{v}^{i}$ & $i$-th level embedding of subtree rooted at node $v$.\\
		$\hat{f}_{v}^{(t)}$ & Embedding of the subtree rooted at node $v$ with $t$-levels.\\
		$\mathbf{\hat{F}}_{G}^{(t)}$ & Embedding of the subgraph $G$ generated with the $t$-SE.\\
		
		\hline
		$\chi$ & A set that contains any kind of element.\\
		$X,Y$ & Subsets of $\chi$ with finite elements.\\
		$|X|,|Y|$ & Number of elements in X, Y respectively.\\
		$\varPsi, \psi$ & Kernel function with its corresponding feature map.\\
		
		\hline
		$s$ & Solid function defined for the elements. \\
		$\varPhi$ & Set of all matching schemes induced by injections.\\
		$\varOmega^{*}$ & Optimal match for two sets.\\
		$k_{e}$ & Element kernel defined for the elements.\\
		$k_{s}$ & Set kernel defined for the sets.\\
		$ s_{i,j} $ & Similarity between $i$ and $j$. \\
		$ S_{i,j} $ & Element in the $i$-th row and $j$-th column of $\mathbf{S}$. \\
		$ k_{\sqcap} $ & Histogram intersection kernel defined for vectors.\\
		$ V_{in}(T) $ & All the internal nodes of the hierarchical tree $T$.\\
		\hline 
	\end{tabular}
\end{table}

\section{Problem Formulation and Preliminaries}
\subsection{\textbf{Problem Formulation}}
\noindent {Definition 1 (\textit{Graph}):} Let $G=(V_{G}, E_{G})$ represents an undirected graph, where $V_{G}=\left\lbrace v_{1},v_{2},\dots v_{n} \right\rbrace $ denotes the set of nodes and $E_{G}=\left\lbrace e_{1},e_{2},\dots e_{m} \right\rbrace $ denotes the set of edges. $\mathbf{A}_{G}\in\mathbb{R}^{n\times n}$ is the adjacency matrix recording the connection information between $n$ nodes. $\mathbf{F}_{G}\in\mathbb{R}^{n\times d}$ is the feature matrix describing the $d$-dimensional node features. $\mathcal{N}(u)$ represents the set of one-hop neighbors for node $u$, while $G_{u}^{k}$ denotes the subgraph formed by the nodes centered on $u$ within $k$ hops.

\noindent {Definition 2 (\textit{Disentangled Representation}):} Given the subgraph $G_{u}^{k}$ and trainable graph filters $\gamma=\{\gamma_{i}|1\leq i\leq T\}$, the disentangled representation $\mathbf{z}_{u}$ of node $u$ is defined as the similarities between its subgraph and the individual graph filters.
\begin{equation}
	\label{feature_update}
	\begin{split} 
		\mathbf{z}_{u} = [\kappa(G_{u}^{k},\gamma_{1}), \kappa(G_{u}^{k},\gamma_{2}), \dots,\kappa(G_{u}^{k},\gamma_{T})],
	\end{split}
\end{equation}
where $\kappa$ is a graph kernel used to measure the similarity between subgraphs and graph filters.

Based on definitions, the main obstacle to solving is designing a differentiable, accurate, and interpretable graph kernel for graph pattern recognition, thereby enhancing the quality of disentangled representation learning for graphs.

Notations frequently used are summarized in Table \ref{tab:notations}.

\subsection{\textbf{Kernel Function}}
A kernel function $\varPsi:\chi \times \chi \rightarrow \mathbb{R}$ can be defined on any non-empty set $\chi$, taking two items as input and quantifying their similarity. $\varPsi$ transforms the items into vectors in a high-dimensional Hilbert space $\mathcal{H}$ and outputs the inner product. Each kernel function uniquely corresponds to a specific feature map $\psi$. The kernel function $\varPsi$ must be clearly defined, while the feature map $\psi$ is often neither feasible nor required to be explicitly expressed. For $\forall x,y\in \chi$, the kernel value can be expressed as:
\begin{equation} 
	\varPsi(x,y) = \left\langle \psi (x), \psi (y) \right\rangle_{\mathcal{H}}.
\end{equation}

\subsection{\textbf{$R$-Convolution Framework}}
$R$-convolution framework provides a unified kernel method for measuring graph similarity. It decomposes two graphs into sets of substructures and measures the graph similarity by the sum of similarities between all-substructure-pairs, as shown in Fig.\ref{R_Convolution}. The $R$-convolution framework adopts an open definition for the decomposition of graphs, e.g., nodes, subgraphs, and random walks:
\begin{equation} 
	\kappa(G,H) = \sum_{g_{i}\in R^{-1}(G)} \sum_{h_{i}\in R^{-1}(H)} \kappa_{base}(g_{i},h_{i}).
\end{equation}
where $R^{-1}$ represents the set of substructures obtained by decomposing a graph, and $\kappa_{base}$ is a kernel function used to measure the similarity between substructures.

\subsection{\textbf{Label Refinement}}
\begin{figure}[t]
	\centering
	\subfloat[R-Convolution]{\includegraphics[height=1.25in]{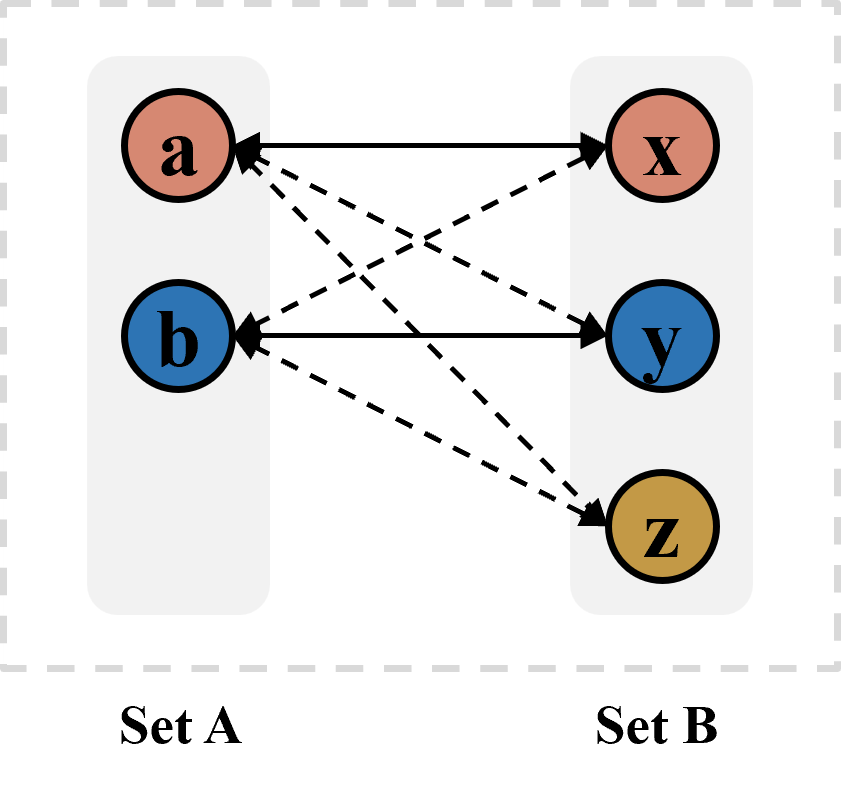}%
		\label{R_Convolution}}
	\hfil
	\subfloat[Label Refinement]{
		\includegraphics[height=1.25in]{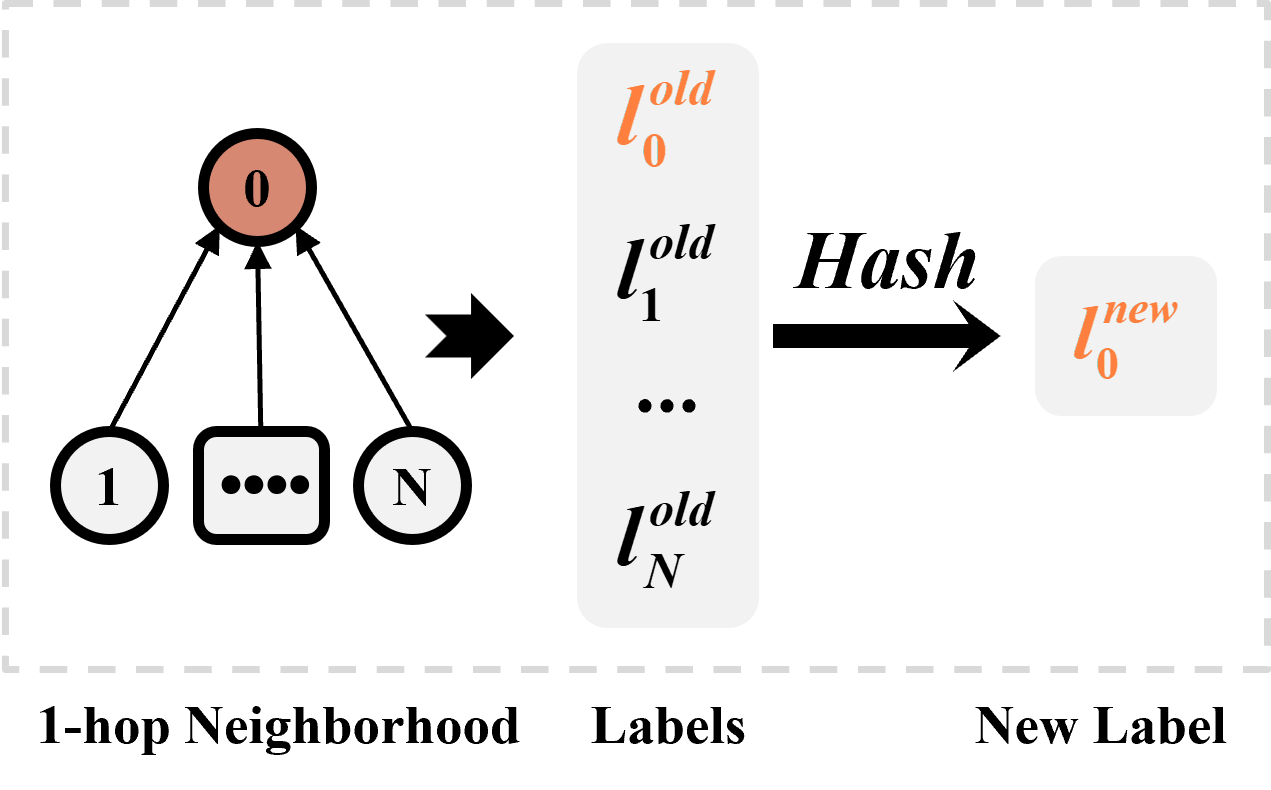}%
		\label{hash}}
	\caption{Key mechanisms in graph kernel. (a) The $R$-Convolution framework calculates the graph similarity by the sum of similarities between all-substructure pairs. However, what matters may be the similarities between matched substructures, denoted by solid lines, with the dashed lines representing the spurious similarities. (b) The Label Refinement Mechanism seeks to encode a tree by mapping all node labels to a new label in an injective manner. Nevertheless, when confronted with various unknown trees, a definitive and differentiable hash function remains to be found.}
	\label{kernel}
\end{figure}

The Label Refinement mechanism updates a node's label by encoding its and neighbors' labels into a new one, as shown in Fig.\ref{hash}. This new label uniquely characterizes the node's neighborhood. Through multiple iterations of Label Refinement on all nodes in the graph, each new label captures the receptive field of the node's multi-hop neighbors.
\begin{equation} 
	l_{v}^{new} = f(l_{v}^{old}, g(\left\lbrace l_{u}^{old}|u\in\mathcal{N}(v)\right\rbrace )).
\end{equation}

To ensure the expressive power of node labels, both $g$ and $f$ should be injective functions. Additionally, $g$ must satisfy permutation invariance.

\subsection{\textbf{Graph Neural Networks}}
Graph neural networks learn a representation for each node through a differentiable iterative message-passing mechanism. They adhere to the same paradigm: updating a node's features by iteratively aggregating features from itself and its neighbors:
\begin{equation}
	\label{gnn}
	\begin{split} 
		\mathbf{h}_{u}^{i+1} = Combine(\mathbf{h}_{u}^{i},Aggregate(\mathbf{h}_{v}^{i}|v\in\mathcal{N}(u))).
	\end{split}
\end{equation}

In the combination process, the node features are transformed into a more discriminative feature space through a set of learnable vector-based filters coupled with an activation function. The neighbor aggregation mechanism can be regarded as a generalized Label Refinement that enforces differentiability at the cost of losing injectivity.

\subsection{\textbf{Expressive Power of GNNs}}
The expressive power of a graph neural network refers to its ability to encode and process the information of node features and topological structures in the graph. This capability can be characterized as subgraph counting and separation ability in node-level and graph-level embeddings, respectively\cite{zhang2024expressive}. It reflects the model's capability to generate distinct embeddings for different subgraphs or graphs.

\section{Graph Optimal Matching Kernel Convolutional Network}
\begin{figure*}[t]
	\centering
	\includegraphics[width=7in]{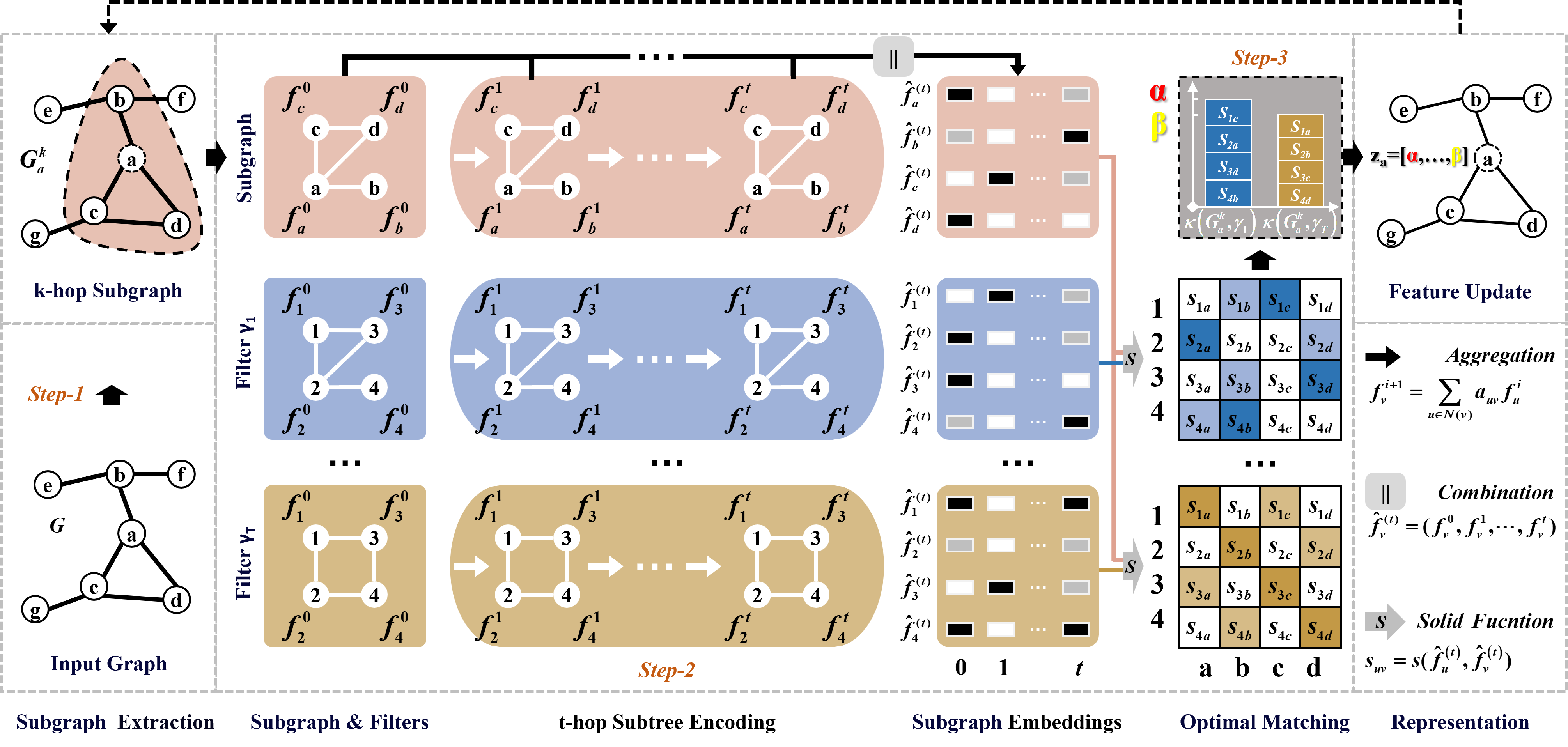}
	\caption{The framework of the Graph Optimal Matching Kernel Convolutional Network. For an input graph, the subgraphs centered around each node within $k$ hops are fed into the model for pattern analysis. Each subgraph is compared with the graph filters to generate the disentangled node representation. The model iteratively updates node features in the graphs using the neighbor aggregation mechanism. Concatenating the historical features of a node forms a multi-resolution representation of the subtree rooted at this node. Each subgraph or filter embedding is denoted by the set of subtree embeddings. Structural alignment between subgraphs and graph filters can be accomplished based on the subtree embeddings with the optimal match kernel. The similarity between graphs is measured as the sum of paired local structural similarities.}
	\label{framework}
\end{figure*}
This section presents the proposed structure-oriented disentangled graph representation learning method, GOMKCN, as illustrated in Fig.\ref{framework}.
\subsection{\textbf{Framework}}
The GOMKCN performs convolution on the graph to generate a disentangled representation for individual node-centric subgraphs. In particular, a novel graph kernel named GOMK is proposed to serve as the convolution operator. It compares the subgraph and a series of trainable graph filters within the model, utilizing the similarities as the disentangled representation of the subgraph. Overall, the working process is divided into three stages: subgraph extraction, subgraph embedding, and subgraph matching. Notably, the edge weights and node features of graph filters are both learnable parameters, aiming to capture the critical structural patterns within the graph.

\noindent {\bf{Step-1: Subgraph Extraction.}} The input graph is divided into subgraphs centered on individual nodes within $k$ hops. The subgraph centered on $v$ can be expressed as:
\begin{equation}
	\label{subgraph}
	\begin{split} 
		G_{u}^{k} &= (V_{G_{u}^{k}}, E_{G_{u}^{k}}) \\
		where\ \ \ \ \ \ \ \ \ \ \ \ \ \ V_{G_{u}^{k}} &= \left\lbrace v\in V_{G}|d(u,v)\leq k\right\rbrace \\
		and\ \ \ \ \ \ \ \ \ \ \ \ \ \ \ \ \ E_{G_{u}^{k}} &= \left\lbrace(u,v)\in E_{G}|u,v\in V_{G_{u}^{k}}\right\rbrace,
	\end{split}
\end{equation}
where $V_{G_{u}^{k}}$ represents the set of $k$-hop reachable nodes from node $u$ in graph $G$ and $E_{G_{u}^{k}}$ represents the edges in $G$ that connect nodes within $V_{G_{u}^{k}}$.

Since the subgraphs encapsulate all information about each node and describe the details of the graph at various localizations, such a decomposition rarely causes loss of information. By projecting each subgraph onto a point in Hilbert space, the whole graph can be characterized as a set of such points. Subsequently, each subgraph will be input into the model to produce the decoupled representation by calculating the similarities concerning individual graph filters. Mathematically, the similarities derived from GOMK correspond to the projections of the subgraph onto individual graph filters within the Hilbert space.

\noindent {\bf{Step-2: Subgraph Embedding.}} Inspired by the $R$-convolution framework, we quantify the similarity between subgraphs and graph filters by measuring the similarity of their subtrees. In particular, each subgraph or graph filter is further decomposed into a set of $t$-level subtrees rooted at individual nodes. To measure the similarity between subtrees, a $t$-SE method is introduced to produce the embedding for each subtree, with the entire subgraph or graph filter being represented by a set of subtree embeddings. Notably, $t$-SE utilizes a summation-based neighbor aggregation mechanism to produce multi-resolution embeddings for subtrees. It satisfies differentiability and operates without the explicit extraction of subtrees. Please refer to Section \ref{subsec:Subgraph_Embedding} for more details.

\noindent {\bf{Step-3: Subgraph Matching.}} After converting the subgraph and the graph filter into sets of subtree embeddings, a novel OMK is proposed to measure the similarity between sets. It establishes an optimal matching between subtrees across the two graphs. Finally, the similarity between the subgraph and the graph filter is measured by aggregating similarities between matched subtree pairs. Based on optimal matching between subtrees, a more accurate and interpretable measure for graph similarity is achieved. Please refer to Section \ref{subsec:Subgraph_Matching} for more details.

\subsection{\textbf{Subgraph Embedding via $t$-level Subtree Encoding}}
\label{subsec:Subgraph_Embedding}
\begin{figure}[t]
	\centering
	\includegraphics[width=3.5in]{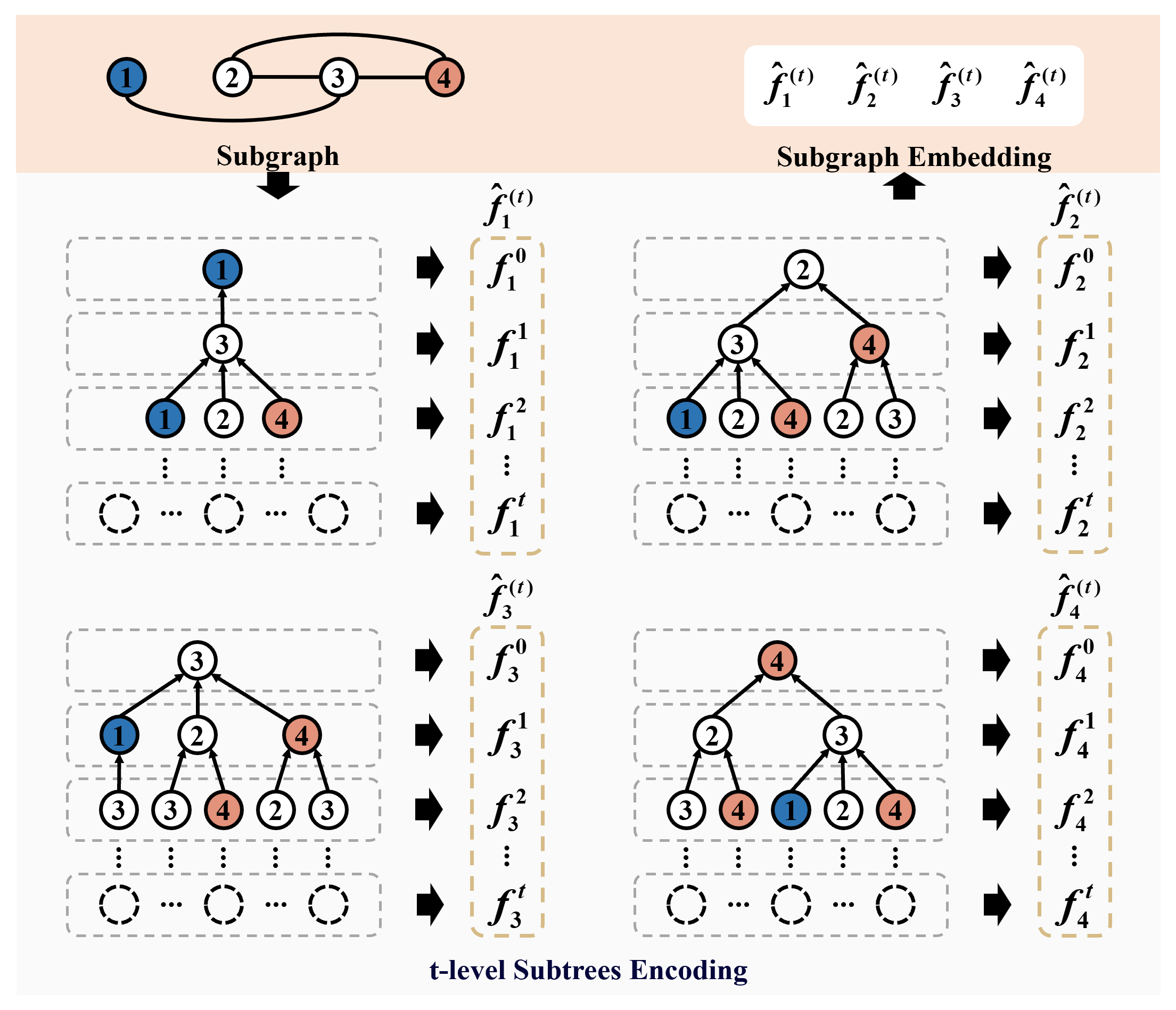}
	\caption{The method of $t$-level Subtree Encoding. Firstly, the subgraph is decomposed into subtrees rooted at individual nodes, noting that the same node often appears multiple times. Secondly, the features of nodes at each level are aggregated towards the root with the weight of edges, respectively, forming layer embeddings. Thirdly, the subtree embedding is defined as the sequence of level embedding, and the set of subtree embeddings forms the subgraph embedding. Importantly, The embeddings can be computed by iteratively multiplying the feature matrix with the adjacency matrix without explicitly extracting the subtrees.}
	\label{subtree}
\end{figure}

Following previous work\cite{shervashidze2011weisfeiler}, we decompose a subgraph or filter into $t$-level subtrees rooted at individual nodes, as shown in Fig.\ref{subtree}. Subsequently, it is natural to generate the embedding for individual subtrees, embedding a subgraph or filter denoted by the set of subtree embeddings. Then, the subgraph and filter similarity can be measured based on their embeddings. Importantly, the generation of embeddings for subgraphs and filters should be as injective as possible to achieve accurate graph pattern recognition. This requirement can be satisfied for finite and fixed subgraphs or filters by encoding each subtree through an iterative Label Refinement, as shown in Fig.\ref{hash}. However, when dealing with unknown graph data and ever-changing filters, no explicit and differentiable algorithm can achieve this goal. Here, we propose the $t$-SE method, which employs a summation-based neighbor aggregation to generate the subtree embeddings. Although the encoding of subtrees may not necessarily be injective, the discovery is that the combination of subtree embeddings exhibits considerable expressive power when representing the subgraph or filter. 

The workflow of $t$-SE is divided into three steps: Level Embedding, Subtree Embedding, and Subgraph Embedding.

\noindent {\bf{1) Level Embedding}:} The features of nodes at the $i$-th layer are aggregated to the root node under the weighting of edges, generating a level embedding $\mathbf{f}_{v}^{i}$, which encodes the structure of $i$-level subtrees and features of nodes in the $i$-th level. Generating such level embeddings can be performed by iterative neighbor aggregation:
\begin{equation}
	\label{level_embedding}
	\begin{split} 
		\mathbf{f}_{v}^{i}= \sum_{u\in \mathcal{N}(v)}a_{uv}\mathbf{f}_{u}^{i-1}.
	\end{split}
\end{equation}

\noindent {\bf{2) Subtree Embedding}:} The layer embeddings are sequentially combined to form the subtree embedding $\hat{f}_{v}^{(t)}$, which provides a multi-resolution representation of the subtree:
\begin{equation}
	\label{subgraph_embedding}
	\begin{split} 
		\hat{f}_{v}^{(t)} = (\mathbf{f}_{v}^{0}, \mathbf{f}_{v}^{1}, \ldots, \mathbf{f}_{v}^{t}).
	\end{split}
\end{equation}
where $\mathbf{f}_{v}^{0}=\mathbf{f}_{v}$ is the original features of node $v$. All subtrees embeddings in $G_{u}^{k}$ can be computed by iteratively multiplying $\mathbf{F}_{G_{u}^{k}}$ with $\mathbf{A}_{G_{u}^{k}}$.

\noindent {\bf{3) Subgraph Embedding}:} The embedding of subgraph centered on $u$ within $k$ hops $\hat{F}_{G_{u}^{k}}^{(t)}$ is defined by the set of subtree embeddings:
\begin{equation}
	\label{Subgraph_embedding}
	\begin{split} 
		\hat{F}_{G_{u}^{k}}^{(t)} = \left\lbrace \hat{f}_{v}^{t}|v\in V_{G_{u}^{k}} \right\rbrace.
	\end{split}
\end{equation}

\subsection{\textbf{Subgraph Matching via Optimal Matching Kernel}}
\label{subsec:Subgraph_Matching}
\begin{figure}[t]
	\centering
	\includegraphics[width=3.5in]{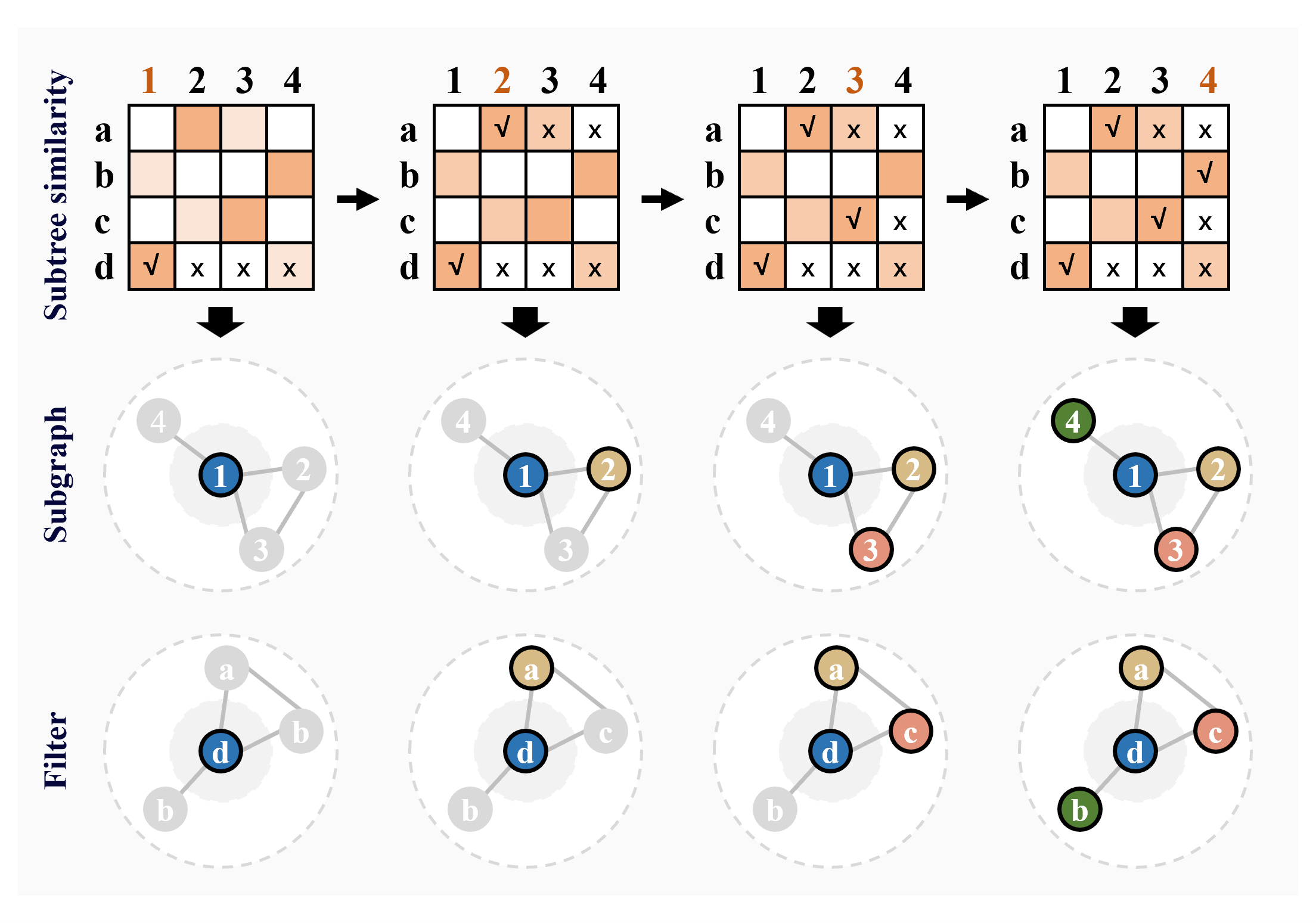}
	\caption{Subtree Matching based on greedy algorithm. Firstly, the similarity between any two subtrees across the subgraph and the filter is calculated by the solid function, as shown in the first row. The darker the color of the square, the higher the similarity between the two subtrees. Secondly, the subtrees within the subgraph sequentially select the most similar counterpart in the filter, as shown in the second and third rows. Once a subtree has been selected, it will no longer participate in subsequent matches. Thirdly, the similarity between the subgraph and the filter is measured by the sum of paired subtrees similarities, indicated by the checkmarks.}
	\label{greedy}
\end{figure}

\begin{figure*}[t]
	\centering
	\subfloat[Gram Matrix]{
		\includegraphics[height=1.35in]{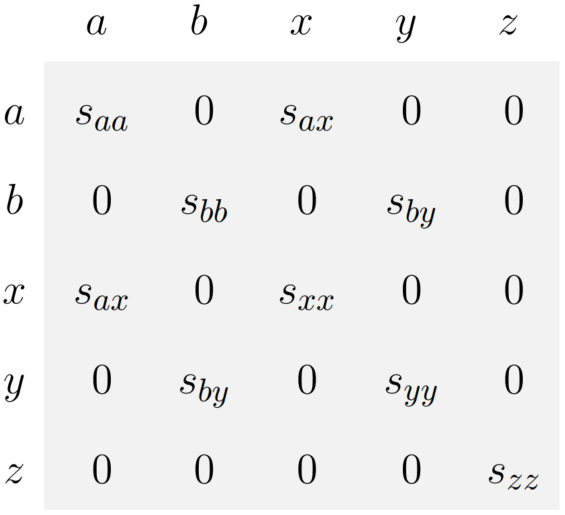}%
		\label{gram_matrix}}
	\hfil
	\subfloat[Hierarchical Tree]{\includegraphics[height=1.35in]{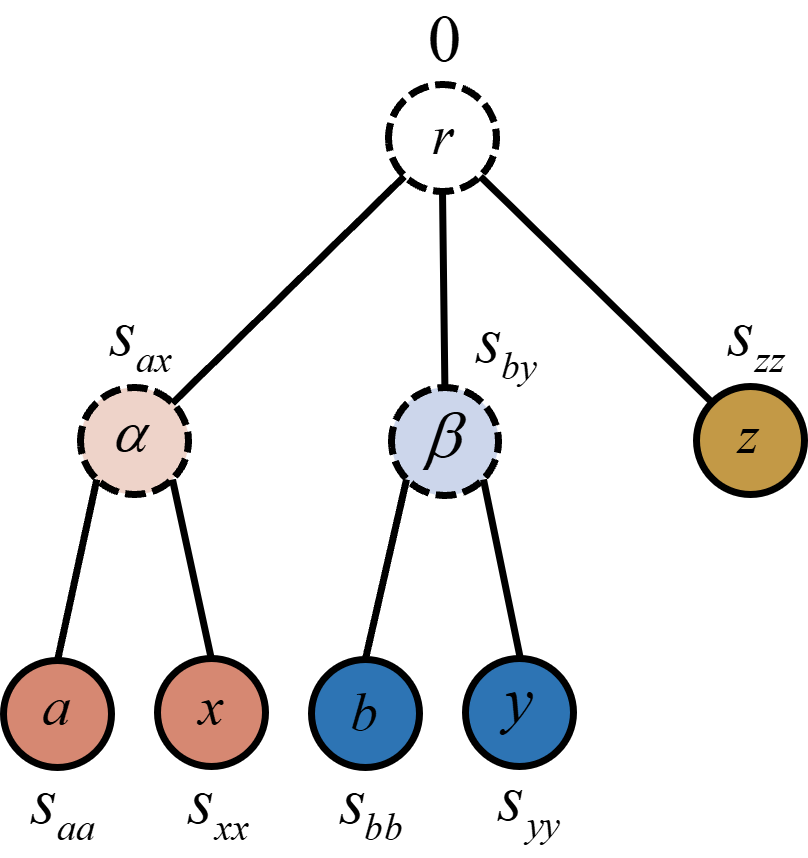}%
		\label{hierarchical_tree}}
	\hfil
	\subfloat[Feature Map]{\includegraphics[height=1.35in]{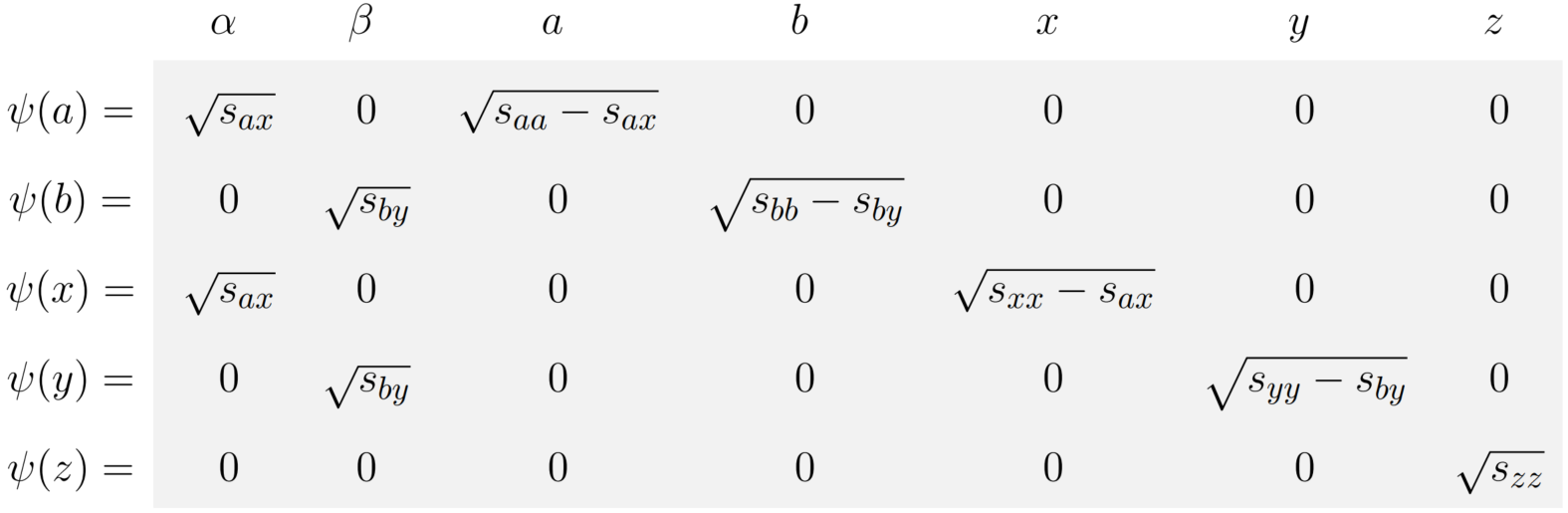}%
		\label{represention}}
	\caption{An example of the Element Kernel $k_{e}$, the elements $a,b$ are members of set $A$, whereas $x,y,z$ belong to set $B$. In particular, $a$ and $b$ are paired with $x$  and $y$, respectively, with $z$ remaining unmatched by any element. (a) The Gram matrix $\mathbf{S}$ describes $k_{e}$ with the items representing the similarities between elements drawn from $A$ and $B$. Similarities between non-matching elements are set to 0 subject to the prior condition of element correspondence. (b) The hierarchical tree $T$ is another representation of $k_{e}$. The similarity between two leaf nodes equals the weight of their lowest common ancestor, $k_{e}(i,j)=\omega (LCA(i,j))$. (c) Feature map $\psi$ of $k_{e}$ can be expressed based on $T$, satisfying $ k_{e}(i,j) = \left\langle \psi (i), \psi (j) \right\rangle $, for $ \forall i,j \in (A\cup B) $.}
	\label{OMK}
\end{figure*}

Traditional graph kernels based on $R$-convolution measure the graph similarity by summing up the similarities of all-substructure pairs, as shown in Fig.\ref{R_Convolution}. Ignoring the correspondence between substructures, these methods may introduce spurious similarities into the graph similarity. To this end, we generalize this question to measure the similarity between two sets and propose the OMK to solve the problem. The workflow of OMK involves four components: element similarity, element matching, set embedding, and set similarity.

\noindent {\bf{1) Element Similarity}:} A solid function needs to be predefined to assess the similarity between any two elements. 

\noindent {Definition 3 (\textit{Solid function}):} For any $x, y\in \chi$, a function $s$ is called solid function if it satisfies $ s:\chi\times\chi\rightarrow\mathbb{R}_{\geqslant 0}, s(x,y) = s(y,x)$ and $s(x,x) \geqslant s(x,y) $.

Given the subtree embeddings of the subgraph and filter in this paper, we instantiate the solid function to measure the subtree similarity. Given any two $t$-level subtrees rooted at $u,v$ respectively, the similarity is defined by Radial Basis Functions (RBFs):
\begin{equation}
	\label{subgraph_similarity}
	\begin{split} 
		s(\hat{f}_{u}^{(t)},\hat{f}_{v}^{(t)}) = \sum_{i=0}^{t}exp(-\dfrac{1}{d\tau}\|\mathbf{f}^{i}_{u}-\mathbf{f}^{i}_{v}\|_{2}^{2}),
	\end{split}
\end{equation}
where $\|\cdot\|_{2}$ is the $L2$-norm, and $\tau$ is a width parameter used to control the matching accuracy.

\noindent {\bf{2) Element Matching}:} It is necessary to align the elements across sets for set similarity. The alignment can be represented by an injective mapping, where each component of the smaller set is mapped to an element in the larger set while allowing some elements in the larger set to remain unmatched.

Different scenarios may involve different matching rules, and one approach is to maximize the overall similarity between all element pairs.

\noindent {Definition 4 (\textit{Optimal Matching}):} The matching that maximizes the sum of similarities between element pairs is called the optimal matching:
\begin{equation}
	\label{}
	\begin{split} 
		\varOmega^{*} = \underset {\varOmega\in\varPhi}{argmax}\sum_{(x,y)\in \varOmega} s(x,y),
	\end{split}
\end{equation}
where $\varPhi$ is the set of all matching schemes and $\varOmega^{*}$ is the optimal matching.

The solution for $\varOmega^{*}$ belongs to combinatorial optimization problems. Although the Kuhn-Munkres algorithm\cite{kuhn1955hungarian},\cite{munkres1957algorithms} is well-suited to find the optimal matching. However, current machine-learning platforms lack parallel optimization for this algorithm. In this paper, we propose a greedy algorithm as an alternative solution for subtree matching: given a subgraph and a filter, each represented by a collection of subtrees, the algorithm sequentially seeks the optimal matching subtree for the subtrees in the subgraph, with already matched subtrees being excluded. A toy example of the working flow is illustrated in Fig.\ref{greedy}. Although such a greedy algorithm may find a suboptimal matching solution, it enables parallel processing of batched data and achieves lower time complexity. Moreover, it does not compromise the effectiveness of OMK to serve as a kernel function.

\noindent {\bf{3) Set Embedding}:} Given a solid function and a matching scheme, an element-oriented kernel function can be constructed. It projects all elements to a Hilbert space, and the set embedding can be obtained by summing up the representations of its elements.

\noindent {Proposition 5 (\textit{Element Kernel}):} For $X\in [\chi]^{m},Y\in [\chi]^{n},m\leq n,x\in X,y\in Y$, and $\varOmega$ is the matching scheme, the function $k_{e}$ induced by only retaining the similarity between matched elements and the similarity of an element to itself is a valid kernel function. 
\begin{equation}
	\label{element_kernel}
	{k_{e}(x,y)} = \begin{cases}
		s(x,y),&{\text{if}}\ x=y\ or\ (x,y) \in \varOmega^{*} \\ 
		{0,}&{\text{otherwise.}} 
	\end{cases}
\end{equation}

\noindent \textit{proof:}

According to Moore–Aronszajn theorem\cite{aronszajn1950theory}, a function is a valid kernel function if its Gram Matrix satisfies symmetry and positive semi-definiteness. The element kernel $k_{e}$ can be described by a Gram Matrix $\mathbf{S}$ as Fig.\ref{gram_matrix}. Assuming that the $i$-th element in $X$ is matched with the $i$-th element in $ Y $ by rearranging the order of elements. It is straightforward to demonstrate that Gram Matrix $\mathbf{S}$ is symmetric. For $\forall\mathbf{\xi}\in \mathbb{R}^{m+n}$, the quadratic form of Gram Matrix $S$ is non-negative:
\begin{equation*}
	\label{local_kernel}
	\begin{split} 
		\mathbf{\xi^{T} S \xi} &=\sum_{i=1}^{m+n} \sum_{j=1}^{m+n} \xi_{i}\xi_{j}S_{i,j}\\
		&=\sum_{i=1}^{m+n} \xi_{i}^{2}S_{i,i} + 2\sum_{i=1}^{m} \xi_{i}\xi_{i+m}S_{i,i+m}\\
		&\geq \sum_{i=1}^{m} S_{i,i+m}(\dfrac{S_{i,i}\xi_{i}^{2}}{S_{i,i+m}} + 2\xi_{i}\xi_{i+m} + \dfrac{S_{i+m,i+m}\xi_{i+m}^{2}}{S_{i,i+m}})\\
		&\geq \sum_{i=1}^{m} S_{i,i+m}(\xi_{i}+\xi_{i+m})^{2} \geq 0.\\
	\end{split}
\end{equation*}

Given an element kernel $k_{e}$, there exists a unique hierarchical tree $T$ that corresponds to it as shown in Fig.\ref{hierarchical_tree}. The leaf nodes denote elements within all sets. The paired nodes are connected to the same internal node. All internal nodes and leaf nodes representing unmatched elements are connected to the root node $r$. Each node is assigned a weight $\omega$, with the leaf node equal to its self-similarity, the internal node equal to the similarity between its child nodes, and the root node 0. The kernel value between any two elements is equal to the weight of their lowest common ancestor, $k_{e}(i,j) = \omega (LCA(i,j))$.

Based on $T$, the feature map $\psi$ corresponding to $k_{e}$ can be expressed explicitly\cite{kriege2016valid}. For $ \forall \varepsilon\in (X\cup Y) $ , $ k_{e} $ maps $\varepsilon$ into a vector $ \psi(\varepsilon) \rightarrow \mathbb{R}^{2m+n}$ as shown in Fig.\ref{represention}.
\begin{equation}
	[\psi(\varepsilon)]_{v\in V(T)/r} = \begin{cases}
		\sqrt{\omega(v)-\omega(p(v))},&{\text{if}}\ v \in P(\varepsilon) \\ 
		{0,}&{\text{otherwise.}} 
	\end{cases}
\end{equation}
Here, $V(T)/r$ denotes the set of all nodes in $T$ except $r$. $p(v)$ is the parent of node $v$. $P(v)$ represents the set of nodes on the path from $v$ to $r$, including $v$ but excluding $r$.

The set embeddings $\mathbf{\Delta}_{X},\mathbf{\Delta}_{Y}$ can be represented by histograms:
\begin{equation}
	\label{set_embedding}
	\begin{split} 
		\mathbf{\Delta}_{X} = \sum_{\varepsilon \in X} \psi(\varepsilon), \mathbf{\Delta}_{Y} = \sum_{\varepsilon \in Y} \psi(\varepsilon).
	\end{split}
\end{equation}

\noindent {\bf{4) Set Similarity}:} Given the embeddings of two sets, their similarity can be measured by the histogram intersection kernel, which is known to be a valid kernel function.

For two objects $\textbf{g}$ and $\textbf{h}$ represented by vectors, the histogram intersection kernel is defined as  $ k_{\sqcap}(\textbf{g},\textbf{h})=\sum_{i=1}^{d} min(|\textbf{g}_{i}|^{\gamma},|\textbf{h}_{i}|^{\gamma}),\gamma>0$\cite{boughorbel2005generalized}. Here we set $ \gamma=2 $ for simplicity, and the set similarity $k_{s}$ can be calculated:

\begin{equation}
	\label{set_kernel}
	\begin{split} 
		k_{s}(X,Y) &= k_{\sqcap}(\mathbf{\Delta_{X}},\mathbf{\Delta_{Y}})\\
		&= \sum_{i=1}^{2m+n} min\{[\mathbf{\Delta_{X}}]_{i}^{2}, [\mathbf{\Delta_{Y}}]_{i}^{2}\}\\
		&= \sum_{v\in V_{in}(T)} \omega(v)\\
		&=\sum_{(x,y) \in \varOmega} s(x,y),
	\end{split}
\end{equation}
where $k_{\sqcap}$ is the histogram intersection kernel and $V_{in}(T)$ denotes the set consisting of internal nodes in $T$.

The graph optimal matching kernel $\kappa$ can be induced by summing up the similarities between paired subtrees:
\begin{equation} 
	\kappa(G_{u}^{k},\gamma_{i}) = \underset {\varOmega\in\varPhi}{max}\sum_{x\in G_{u}^{k},y\in \gamma_{i},(x,y)\in \varOmega} s(\hat{f}_{x}^{(t)},\hat{f}_{y}^{(t)}).
\end{equation}

\begin{algorithm}[t]
	\caption{GOMKCN.}\label{alg:gomkcn}
	\begin{algorithmic}[1]
		\STATE {\textbf{Input:}} Graph $G=(\mathbf{A}_{G}\in\mathbb{R}^{n\times n},\mathbf{F}_{G}\in\mathbb{R}^{n\times d})$, subgraph radius $k$, iterations of neighbor aggregation $t$, solid function $s$;
		\STATE {\textbf{Output:}} Disentangled representation of subgraphs centered on indivedual nodes $\left\lbrace z_{u}|u\in V_{G}\right\rbrace $;
		\STATE {\textbf{Init:}} Graph filters $\left\lbrace \gamma_{j}|0\leq j\leq T \right\rbrace $;
		
		Step-1 : Subgraph Extraction.
		\STATE {\textbf{for} $u$ in $V_{G}$ \textbf{do}}
		\STATE \hspace{0.5cm} $ G_{u}^{k} \Leftarrow BFS(\mathbf{A}_{G},u)$  //Node-centric subgraph.
		\STATE {\textbf{end}}
		
		Step-2 : Subgraph Embedding.
		\STATE {\textbf{for} $u$ in $V_{G}$ \textbf{or} $j=1,2,\dots,T$ \textbf{do}}
		\STATE \hspace{0.5cm} {\textbf{for} $i=0,1,\dots,t$ \textbf{do}}
		
		\hspace{1.0cm} //Level Embedding for subgraphs and filters.
		\STATE \hspace{1.0cm} $ \left\lbrace f_{v}^{i}|v\in V_{G_{u}^{k}}\right\rbrace  \Leftarrow \mathbf{A}_{G_{u}^{k}}^{i}\mathbf{F}_{G_{u}^{k}}$
		\STATE \hspace{1.0cm} $ \left\lbrace f_{v}^{i}|v\in V_{\gamma_{j}}\right\rbrace  \Leftarrow \mathbf{A}_{\gamma_{j}}^{i}\mathbf{F}_{\gamma_{j}}$
		\STATE \hspace{0.5cm} {\textbf{end}}
		\STATE \hspace{0.5cm} $\hat{f}_{v}^{t} = (\mathbf{f}_{v}^{0}, \mathbf{f}_{v}^{1}, \ldots, \mathbf{f}_{v}^{t})$  //Subtree Embedding.
		\STATE \hspace{0.5cm} $\hat{F}_{G_{u}^{k}}^{(t)} = \left\lbrace \hat{f}_{v}^{t}|v\in V_{G_{u}^{k}} \right\rbrace$  //Subgraph Embedding.
		\STATE \hspace{0.5cm} $\hat{F}_{\gamma_{j}}^{(t)} = \left\lbrace \hat{f}_{v}^{t}|v\in V_{\gamma_{j}} \right\rbrace$  //Filter Embedding.
		\STATE {\textbf{end}}
		
		Step-3 : Subgraph Matching.
		\STATE {\textbf{for} $u$ in $V_{G}$ \textbf{do}}
		\STATE \hspace{0.5cm} {\textbf{for} $j=1,2,\dots,T$ \textbf{do}}
		\STATE \hspace{1.0cm} {$z_{u}^{j}=0$} //Accumulated Subtree Similarities.
		\STATE \hspace{1.0cm} {$\hat{V}=\left\lbrace \right\rbrace$} //Subtrees already matched in the filter.
		\STATE \hspace{1.0cm} {\textbf{for} $v$ in $V_{G_{u}^{k}}$ \textbf{do}}
		\STATE \hspace{1.5cm} //Find the most similar unmatched subtree.
		\STATE \hspace{1.5cm} {$\hat{v}=\underset {v^{'}}{\text{argmax}}\left\lbrace s(\mathbf{f}_{v}^{(t)},\mathbf{f}_{v^{'}}^{(t)})|v^{'}\in V_{\gamma_{j}}/\hat{V}\right\rbrace$}
		\STATE \hspace{1.5cm} {$z_{u}^{j} += s(\mathbf{f}_{v}^{(t)},\mathbf{f}_{\hat{v}}^{(t)})$} //Similarity update.
		\STATE \hspace{1.5cm} {$\hat{V}|=\hat{v}$} //Store the matched subtrees.
		\STATE \hspace{1.0cm} {\textbf{end}}
		\STATE \hspace{0.5cm} {\textbf{end}}
		\STATE \hspace{0.5cm} $z_{u}=[z_{u}^{1},z_{u}^{2},\dots,z_{u}^{T}]$ //Similarities.
		\STATE {\textbf{end}}
		\STATE {\textbf{Return:}} {$\left\lbrace z_{u}|u\in V_{G}\right\rbrace$} //Disentangled representations.
	\end{algorithmic}
\end{algorithm}

Overall, the pseudocode of GOMKCN is listed in Algorithm \ref{alg:gomkcn}. Lines 4-6 present the pseudocode for Step-1, which employs the Breadth-First Search (BFS) algorithm to extract node-centric subgraphs. Lines 7-15 present the pseudocode for Step-2, which utilizes $t$-SE to generate embeddings for individual subgraphs and filters. Lines 16-28 present the pseudocode for Step-3, which employs a greedy algorithm to identify the optimal matching of subtrees between subgraphs and filters, thereby generating disentangled representations. Notably, Step-2 and Step-3 are combined to form the Graph Optimal Matching Kernel $\kappa$.

\section{Property Analysis of GOMKCN}
\subsection{\textbf{Expressive Power of $t$-SE}}
The expressive power of GNNs is fundamentally governed by the degree to which they can satisfy the injectivity requirement for characterizing node neighborhoods or the entire graph. The expressiveness of local neighborhood encodings and global graph embeddings dictates node-level and graph-level prediction performance. They are fundamentally identical, as both constitute representation learning processes for graphs at varying scales, from the local subgraph to the entire graph. GOMKCN explicitly extracts the node-centric subgraphs and transforms each subgraph into a set of subtrees rooted at each node. The $t$-SE employ iterative neighbor aggregation to encode each subtree as a vector embedding and the subgraph is represented by a collection of subtree embeddings. Here, we provide a theoretical analysis of the injectivity guarantees of $t$-SE in subgraph embedding generation.

Given a node-centric subgraph $G_{u}^{k}$ along with its adjacency matrix $\mathbf{A}_{G_{u}^{k}}$ and feature matrix $\mathbf{F}_{G_{u}^{k}}$, $t$-SE performs neighbor aggregation within the subgraph, forming the node embedding with a collection of subtree embeddings:
\begin{equation}
	\begin{split} 
		\mathbf{\hat{F}}_{G_{u}^{k}}^{(t)} = concat\left\lbrace \mathbf{A}_{G_{u}^{k}}^{i}\mathbf{F}_{G_{u}^{k}}|i \in \left[ 0,1,\ldots,t\right] \right\rbrace.
	\end{split}
\end{equation}

Assuming another node $v$ have have identical embeddings with $u$:
\begin{equation}
	\label{gomkcn_embedding}
	\begin{split} 
		\mathbf{A}_{G_{v}^{k}}^{i}\mathbf{F}_{G_{v}^{k}}=\mathbf{A}_{G_{u}^{k}}^{i}\mathbf{F}_{G_{u}^{k}}\ \  where\ \  i=0,1,\ldots,t.
	\end{split}
\end{equation}

\noindent {Proposition 6 (\textit{Expressive power of $t$-SE}):} If there exists a $t$ such that $\mathbf{\hat{F}}_{G_{u}^{k}}^{(t)}=\mathbf{\hat{F}}_{G_{v}^{k}}^{(t)}$ and both are full-rank matrices, then $G_{v}^{k}$ and $G_{v}^{k}$ are isomorphic. 

\noindent \textit{proof:}

Since $\mathbf{\hat{F}}_{G_{u}^{k}}^{(t)}=\mathbf{\hat{F}}_{G_{v}^{k}}^{(t)}$, then $\mathbf{A}_{G_{u}^{k}}^{i}\mathbf{F}_{G_{u}^{k}}=\mathbf{A}_{G_{v}^{k}}^{i}\mathbf{F}_{G_{v}^{k}}=\mathbf{c}_{i}$ holds for $\forall i \in \left[ 0,1,\ldots,t\right]$ where $\mathbf{c}_{i}$ is the introduced constant matrix. Obviously, $\mathbf{F}_{G_{u}^{k}}=\mathbf{F}_{G_{v}^{k}}$ can be obtained by setting $t=0$. What's more, $\mathbf{A}_{G_{u}^{k}}\mathbf{c}_{i}=\mathbf{A}_{G_{v}^{k}}\mathbf{c}_{i}=\mathbf{c}_{i+1}$ holds for $ \forall i \in \left[ 0,1,\ldots,t-1\right] $. For the parameters in the adjacency matrix $\mathbf{A}_{G_{u}^{k}}$ and $\mathbf{A}_{G_{v}^{k}}$, each row comprises a set of $n$ variables, amounting to a total of $n$ such sets. Since $rank(\mathbf{\hat{F}}_{G_{u}^{k}}^{(t)})=rank(\mathbf{\hat{F}}_{G_{v}^{k}}^{(t)})=n$, $n$ inhomogeneous linear equations can be established for each set of variables. Therefore, there exists a unique and identical solution for $\mathbf{A}_{G_{u}^{k}}$ and $\mathbf{A}_{G_{v}^{k}}$. Please note that the above analysis holds regardless of the order of nodes in the matrices.

As derived from the analysis above, increasing the level of subtrees $t$ and the dimension of node features $d$ may enhance $rank(\mathbf{\hat{F}}_{G_{u}^{k}}^{(t)})$, thereby increasing the likelihood of subgraph $G_{u}^{k}$ to possess a unique embedding. But the amount of increase by $t$ is limited, with the maximum being equal to $rank(\mathbf{A}_{G_{u}^{k}})$.

\subsection{\textbf{Spherical Feature Space of GOMK}}
For the GOMK, the similarity between any graph and itself is a constant. This implies that all subgraphs and graph filters are mapped onto the spherical surface with the same radius, regardless of variations in their topological structures and node features. The kernel response of the subgraph and graph filter reflect their cosine similarity in the Hilbert space.

For any subgraph with $n$ nodes, its similarity to itself under the GOMK metric is always a constant, regardless of changes in its topological structure and node features:
\begin{equation}
	\label{}
	\begin{split} 
		\kappa(G_{u}^{k},G_{u}^{k})=\sum_{v\in V_{G_{u}^{k}}} s(\hat{f}_{v}^{(t)},\hat{f}_{v}^{(t)})=n(t+1).
	\end{split}
\end{equation}

This implies that all subgraphs with an equal number of nodes, as well as the filters, are mapped onto the same sphere in the Hilbert space:
\begin{equation}
	\label{}
	\begin{split} 
		\parallel\psi (G_{u}^{k})\parallel=\sqrt{\left\langle \psi (G_{u}^{k}), \psi (G_{u}^{k}) \right\rangle_{\mathcal{H}}}=\sqrt{\kappa(G_{u}^{k},G_{u}^{k})}.
	\end{split}
\end{equation}

Although each node-centric $k$-hop subgraph contains a varying number of nodes, it can be standardized by introducing isolated nodes into the subgraph. Since the original subgraph is a connected graph, a bijection exists between the original subgraph and the adjusted subgraph. Therefore, this operation does not affect the uniqueness of the subgraph representation. As a result, the similarity between the subgraph and filter fully reflects their distance, which is represented by the cosine similarity in Hilbert space. This property significantly facilitates accurate graph pattern learning.
\begin{equation}
	\label{}
	\begin{split} 
		cos(\psi (G_{u}^{k}), \psi (\gamma_{i}))=\dfrac{\left\langle \psi (G_{u}^{k}), \psi (\gamma_{i}) \right\rangle_{\mathcal{H}}}{\parallel\psi (G_{u}^{k})\parallel\parallel\psi (\gamma_{i})\parallel}=\dfrac{\kappa(G_{u}^{k},\gamma_{i})}{n(t+1)}.
	\end{split}
\end{equation}

\section{Experiments}
This chapter experimentally evaluates the proposed GOMKCN with respect to the following questions:
\begin{itemize}
	\item{\textit{Q1.} Can the model injectively embed the subgraphs and accurately learn the target pattern?}(Section \ref{subsec:Expressive_Power})
	\item{\textit{Q2.} Can the model effectively discover the frequent subgraph patterns in the graph?}(Section \ref{subsec:Frequent_Pattern})
	\item{\textit{Q3.} How is the interpretability of the model reflected during the prediction process?}(Section \ref{subsec:Interpretability_Analysis})
	\item{\textit{Q4.} How is the prediction performance of the model on real-world datasets?}(Section \ref{subsec:Node_Classification}, \ref{subsec:Graph_Classification})
	\item{\textit{Q5.} How do the model's hyperparameters affect its performance?}(Section \ref{subsec:Hyperparameter_Analysis})
	\item{\textit{Q6.} What operational time complexity does the model exhibit empirically?}(Section \ref{subsec:Time_Complexity})
\end{itemize}

\subsection{Expressive Power}
\label{subsec:Expressive_Power}

\begin{figure*}[t]
	\centering
	\subfloat[Framework of Isomorphic Graph Learning]{
		\includegraphics[height=2.3in]{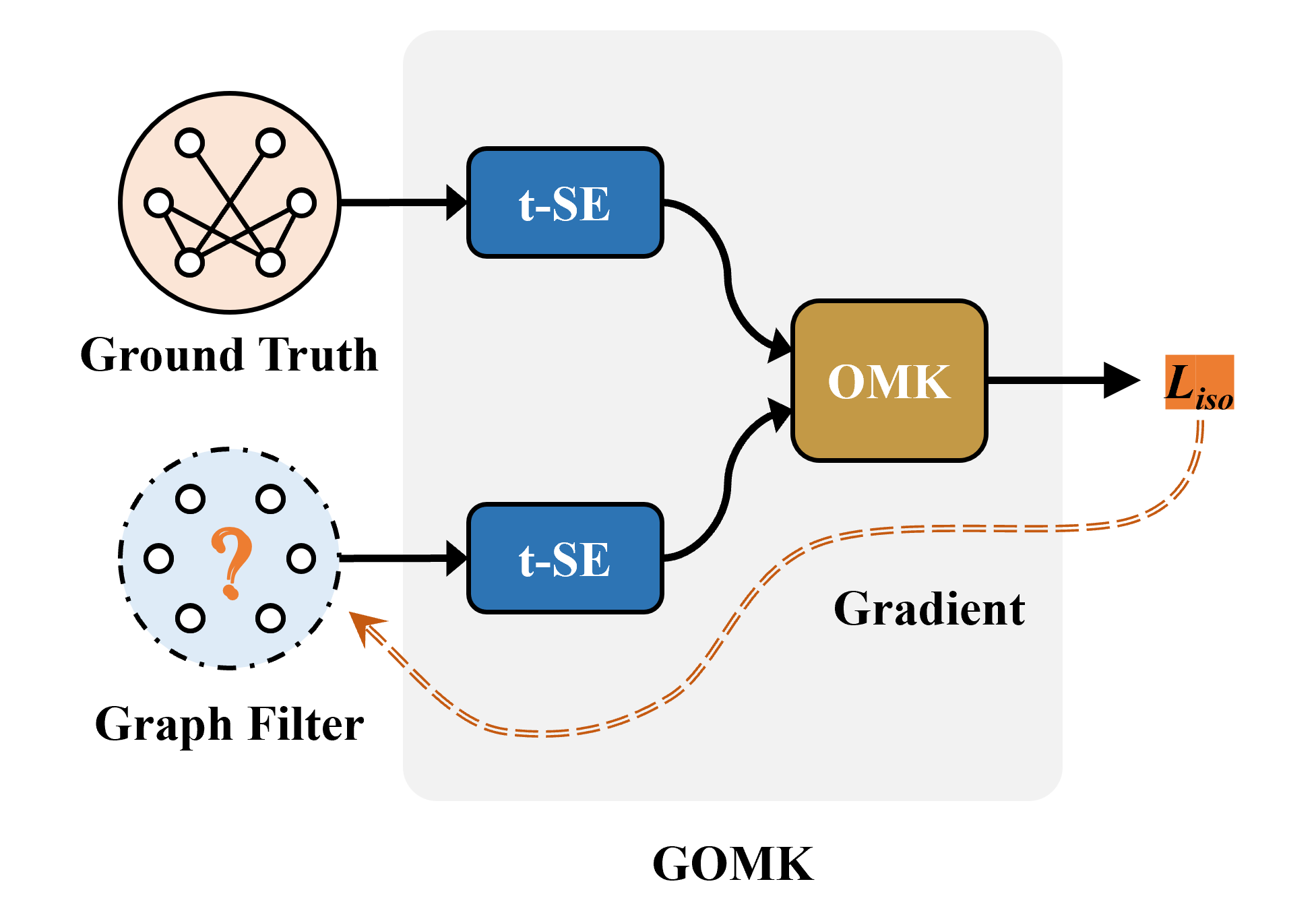}%
		\label{iso_framework}}
	\hfil
	\subfloat[Variation of Graph Filter]{\includegraphics[height=2.3in]{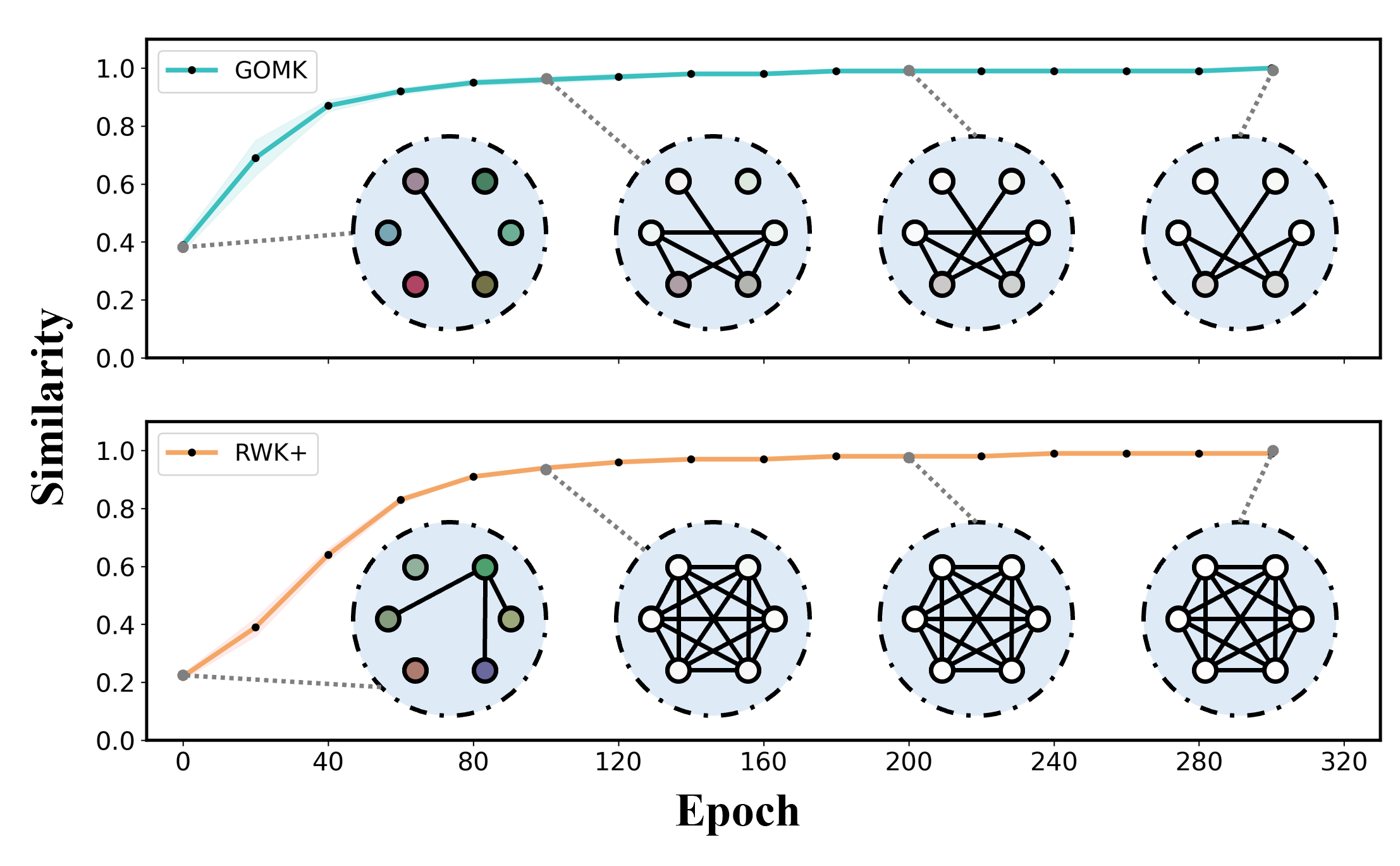}%
		\label{iso_filter}}
	\hfil
	\subfloat[Results of Isomorphic Graph Learning]{\includegraphics[width=7in]{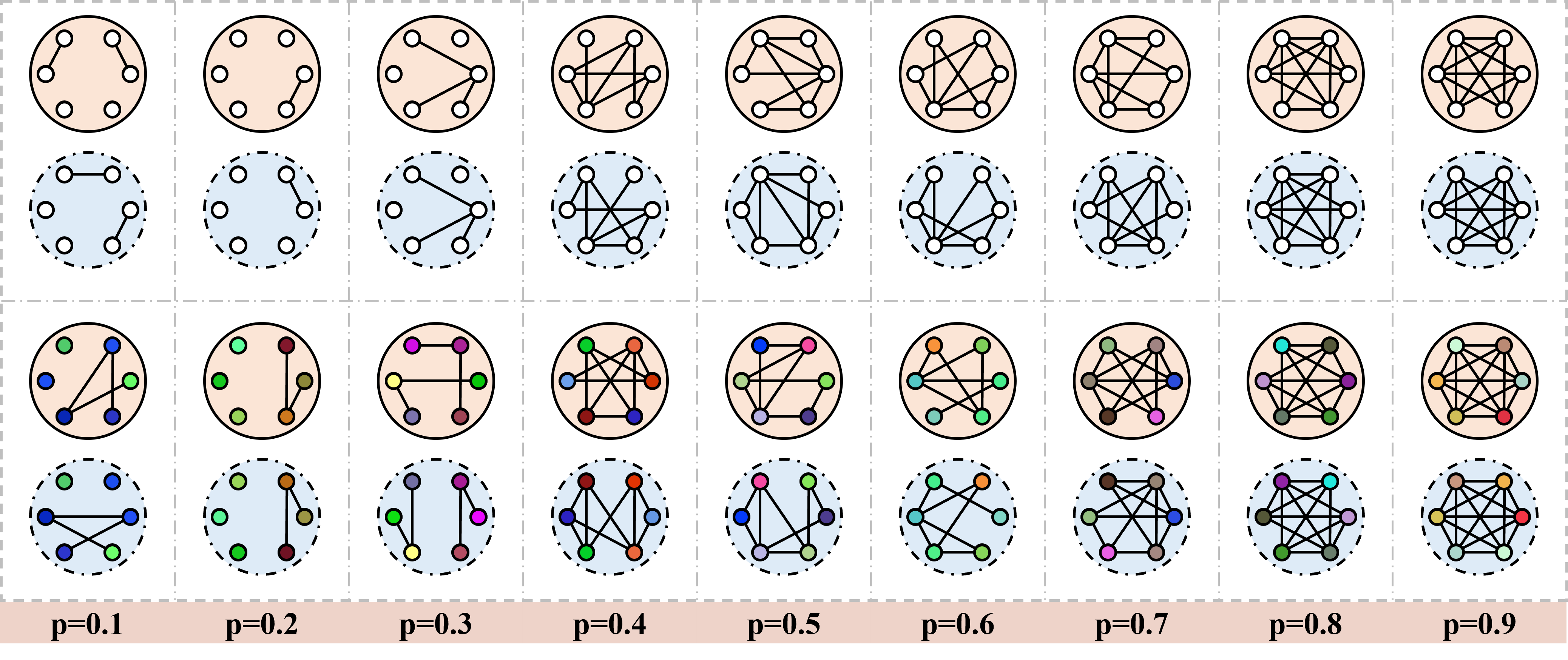}%
		\label{iso_result}}
	\caption{The framework and outcomes of isomorphic graph learning. (a) The framework of isomorphic graph learning. The ground truth and the graph filter are input into GOMK for comparison, and the graph filter will be updated repeatedly to maximize their similarity. Whether the graph filter is consistent with the ground truth after training reflects its ability to identify graph patterns accurately. (b) A comparison between RWK$^{+}$CN and GOMK in isomorphic graph learning. Graph filters are continuously updated as the similarity increases. (c) Results of isomorphic graph learning. The 1st and 3rd rows represent the ground truth when the node features are set to all ones and random numbers, respectively, while the 2nd and 4th rows represent the training results of the graph filter when given the corresponding ground truth.}
	\label{ISO}
\end{figure*}

In this Subsection, we conduct an experiment called Isomorphic Graph Learning to validate the model's expressive power. We focus on verifying whether the GOMK helps to learn a precise graph in the filter consistent with a given graph pattern. This is crucial for determining whether the model can accurately capture graph patterns.

\noindent {\bf{Method.}} The framework of Isomorphic Graph Learning is illustrated in Fig.\ref{iso_framework}. We feed a randomly generated graph $G$, serving as ground truth, and a trainable graph filter $H$ into the GOMK $\kappa$. The output of GOMK reflects the similarity between the ground truth and the graph filter. After training the model by maximizing the similarity, we verify whether the graph filter is consistent with the ground truth, including the topological structure and the node features.
\begin{equation}
	\label{iso_graph}
	\begin{split} 
		L_{iso}=-\underset {H}{\text{argmax}}{\kappa(G,H)}.
	\end{split}
\end{equation}

\noindent {\bf{Settings.}} Here, we only consider undirected graphs with six nodes. The edge weights are either 0 or 1, and each node has three-dimensional features. The graph filter has the same shape as the ground truth, with the adjacency matrix and the feature matrix being trainable parameters. In particular, all these parameters are bounded within the [0,1] range. We utilize randomly generated graphs to act as the ground truth respectively. The connection between nodes follows a Bernoulli distribution $A_{ij}=A_{ji}\sim Bernoulli(p)$. By varying the value of $p$ within [0.1, \dots, 0.9], both sparse and dense graphs can be generated. Both scenarios, whether nodes have features or not, are considered. When solely considering the graph structure, all node features are initialized as 1.0; otherwise, they are randomly sampled from a uniform distribution over [0,1]. There are three steps for neighbor aggregation, and the width parameter is set to 1.0. The model is trained for 500 epochs with a learning rate of 0.5.

\noindent {\bf{Results.}} We randomly selected a graph as the ground truth to demonstrate the variation of the graph filter during the training process. In particular, the model is trained for 300 epochs with a learning rate 0.05 for clear visualization. We compared the results of the GOMK with the improved random walk kernel in RWK$^{+}$CN\cite{lee2024descriptive}. The ground truth is illustrated in Fig.\ref{iso_framework}, and the variation of the graph filter is shown in Fig.\ref{iso_filter}. An edge between nodes is displayed if the corresponding weight in the adjacency matrix is greater than 0.5; otherwise, it is not shown. The node features were transformed into RGB values corresponding to 1 pixel for visualization. It is found that GOMK achieves accurate learning in both topology and features, whereas RWK$^{+}$ tends to learn a network with a fully connected topology and features all set to 1. This may be attributed to the oscillation phenomenon in the random walk kernel, which allows nodes to be visited multiple times, leading to false similarities between graphs. Additionally, We report more training results of the graph filter with different random graphs serving as the ground truth, as shown in Fig.\ref{iso_result}. Experimental results show that GOMK accurately learns graph topological structures and node features, even for graphs with low-rank adjacency matrices. This implies that when restricted to unweighted and undirected graphs with node features bounded in [0,1], $t$-SE generates injective embedding for these graphs.

\subsection{Frequent Subgraph Pattern Mining}
\label{subsec:Frequent_Pattern}
Frequent pattern mining aims to discover the most common communities in a graph. In practical applications, the topological structure, node features, and occurrence frequency of patterns are all important subjects. The GOMKCN helps to accurate and efficient pattern analysis of graph data. Importantly, it simultaneously learns multiple frequent patterns in the graph rather than just the most frequent one.

\begin{figure}[t]
	\centering
	\subfloat[Synthetic Graph]{
		\includegraphics[height=1.5in]{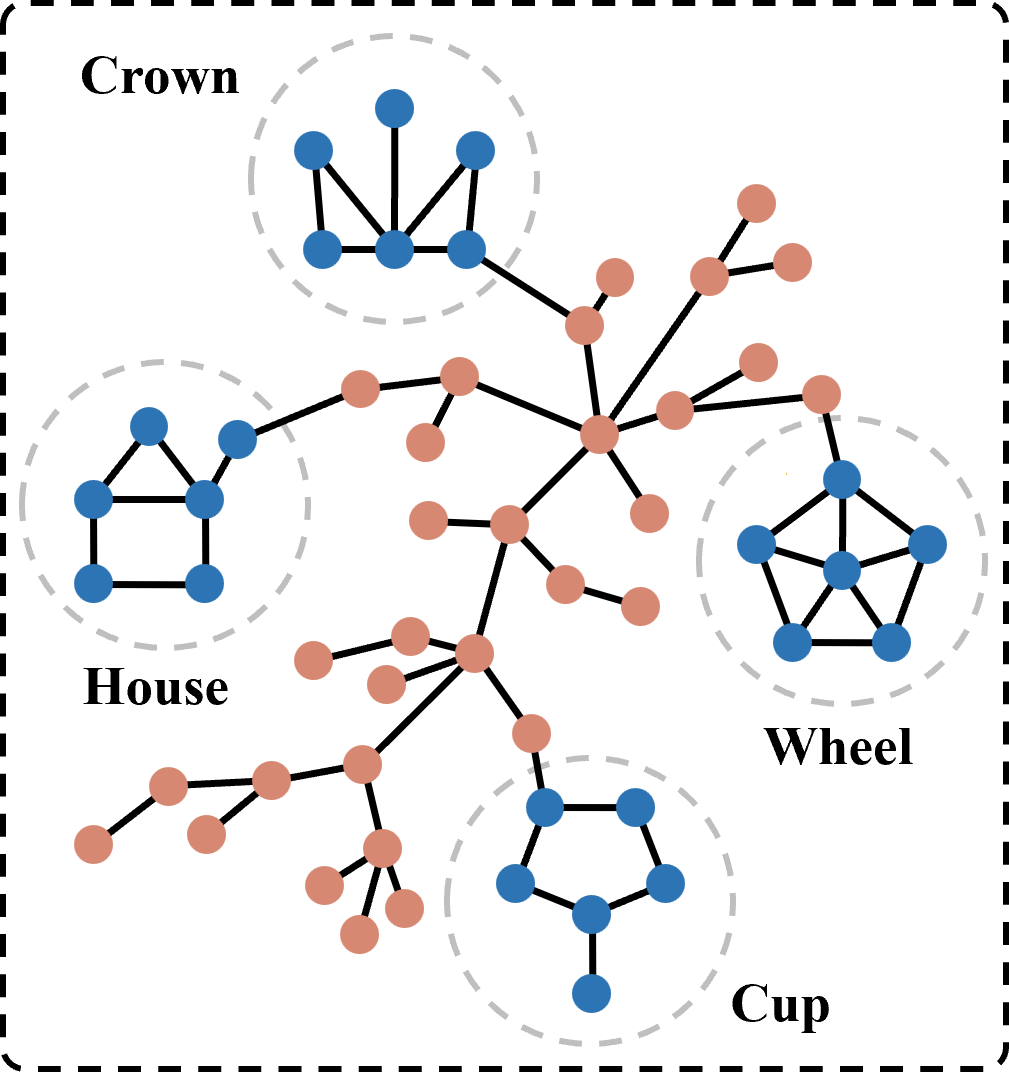}%
		\label{synthetic_graph}}
	\hfil
	\subfloat[Graph Filters]{
		\includegraphics[height=1.5in]{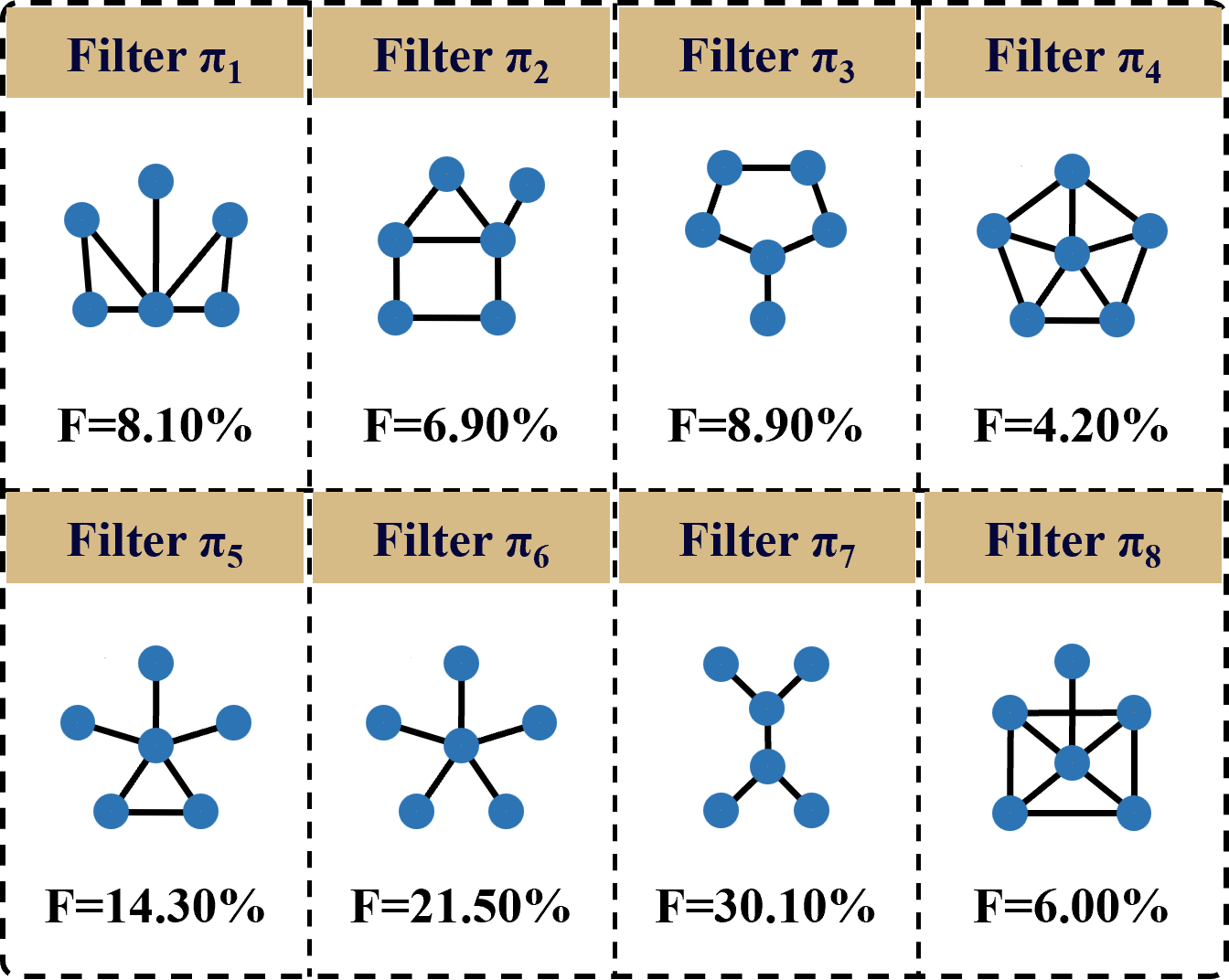}%
		\label{graph_filters}}
	\caption{Illustration of frequent graph pattern mining. (a) Synthetic graph using the Barabási–Albert random graph and the motifs. (b) Graph patterns with their frequencies of occurrence discovered by the graph filters.}
	\label{patterns}
\end{figure}

\noindent {\bf{Dataset.}} We constructed an artificial graph containing 1000 nodes by incorporating four types of motifs into a Barabási–Albert (BA) random graph: House, Cup, Wheel, and Crown. The BA graph has a total of 760 nodes, and the degree of nodes follows a power-law distribution. Besides, any motif is attached to the BA graph through a randomly selected node from each graph, respectively. The quantity of each type of motif is set to 10. The nodes in both the BA graph and the motif have 3-dimensional features randomly initialized by a uniform distribution over the interval [0,1]. The diagram of the dataset is shown in Fig.\ref{synthetic_graph}.

\noindent {\bf{Method.}} For the subgraphs centered on each node, identifying frequent patterns belongs to clustering problems. This task can be accomplished by the GOMKCN, which has trainable graph filters that serve as clustering centers. Suppose the $k$-hop subgraphs are extracted on $p$ nodes. The number of graph filters is set to $q$, each with the same number of nodes as the subgraphs. In this way, we aim to discover the $q$ most frequent patterns among $p$ subgraphs. GOMKCN calculates the similarity between each subgraph and each graph filter. Since all graphs, including the graph filters with learnable parameters, are mapped onto the same spherical surface centered at the origin in Hilbert space, the kernel response is proportional to their cosine similarities. Like the k-means algorithm, we define the training loss as maximizing the similarity between each subgraph and its corresponding frequent subgraph.
\begin{equation}
	\label{frq_pattern}
	\begin{split}
		L_{frq}=-\underset {\gamma_{1},\gamma_{2},\dots,\gamma_{q}} {argmax} \sum_{i=1}^{p} \underset{j\in\{1,2,\dots,q\}} {max}{\kappa(G_{i}^{k},\gamma_{j})}.
	\end{split}
\end{equation}

\noindent {\bf{Settings.}} The subgraphs centered on each node within a 3-hop distance in the graph are fed into the model to mine frequent patterns. The number and size of graph filters in GOMKCN are set to 8 and 6, respectively. The steps for neighbor aggregation are set to 3. The model is trained for 500 epochs with a learning rate of 0.5.

\noindent {\bf{Results.}} After the model converges, the graph filters demonstrate the discovered frequent patterns, as shown in Fig.\ref{graph_filters}. Among the graph patterns, filters $\gamma_{1}-\gamma_{4}$ accurately reveal the motifs that have been manually embedded, while $\gamma_{5}-\gamma_{7}$ seems randomly generated patterns in the BA graph and $\gamma_{8}$ looks like a partial motif of $\gamma_{4}$. The reason is that the subgraph extraction process for motif nodes incorporates nodes from the BA graph but excludes certain motif nodes. By assigning each subgraph to its most similar graph filter, we have approximately estimated the frequencies of individual patterns.

\subsection{Interpretability Analysis}
\label{subsec:Interpretability_Analysis}
\begin{figure*}[t]
	\centering
	\includegraphics[width=7.1in]{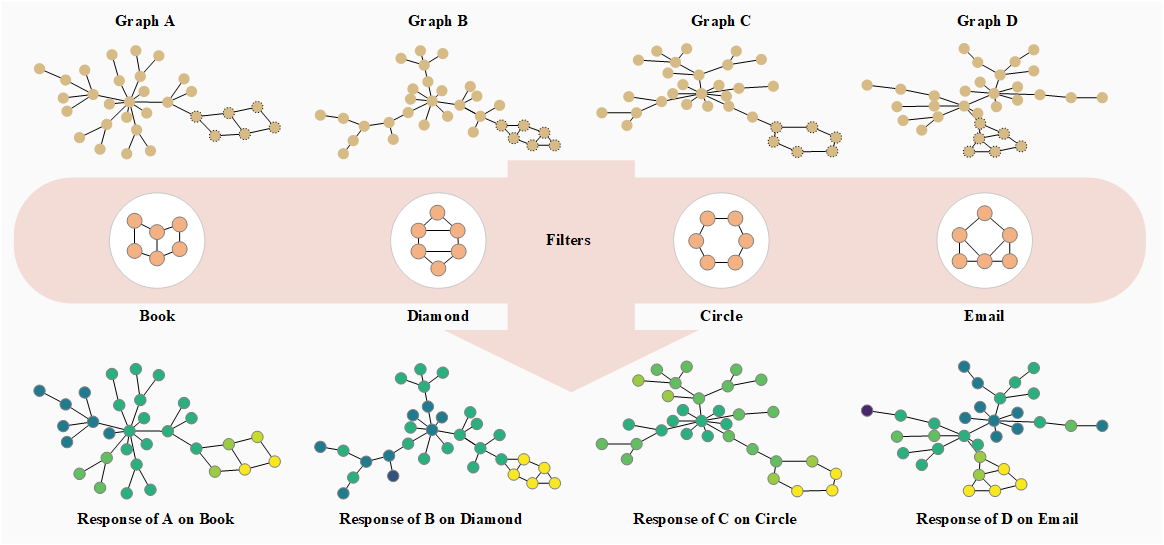}
	\caption{Illustration of model interpretability. The first row displays four graph samples in the test set, each belonging to a distinct category. Each graph comprises a Barabási–Albert random graph and a motif. The type of motif serves as the ground truth, determining the label of the entire graph. The second row displays the graph filters after the model has been trained. The third row shows the kernel response of node-centric subgraphs on the corresponding graph filters. The color of a node indicates the kernel response of its subgraph's similarity to the corresponding graph filter. The color intensity correlates with the strength of response: yellow indicates high response, while blue signifies low response.}
	\label{response} 
\end{figure*}

In GOMKCN, graph filters serve as the foundation upon which the model relies for inference. This implies that under the guidance of loss functions in classification tasks, graph filters will capture the common characteristics among graphs within the same category and the distinguishing features between various categories. Thus, they provide a powerful explanation for revealing the model's reasoning process. 

\noindent {\bf{Dataset.}} To validate the model's accuracy in graph pattern mining, we construct an artificial dataset comprising 8000 graphs. Each graph consists of a Barabási–Albert (BA) Random graph with 25 nodes and one of the motifs: Book, Diamond, Circle, or Email. The graphs are classified into four distinct categories, with the label of any graph being determined by the specific type of motif it contains. So, these motifs act as ground truth for graph classification. The nodes in the BA graph and the motifs have 3-dimensional features, which are initialized using a uniform distribution over [0,1]. The dataset is divided into training, validation, and test sets in the proportion of 8:1:1. Four samples were extracted from the test set and displayed in the 1-th line of Fig.\ref{response}, each belonging to different categories of graphs.

\noindent {\bf{Settings.}} We employ GOMKCN with four graph filters for classification. Each filter has six nodes, and the dimension of the node feature is 3. The model is trained for 200 epochs with a learning rate of 0.1. The steps for neighbor aggregation are set to 2, and the width parameter is set to 0.5. Every 512 graphs form a batch for training, and the 3-hop subgraphs of each node are fed into the model for pattern recognition. Subsequently, we apply max pooling to aggregate node representations into graph-level representations.

\noindent {\bf{Results.}} After the training, the model achieved perfectly accurate predictions on the test set of input graphs. The graph filters are illustrated in the 2nd line of Fig.\ref{response}. Furthermore, we visualize the kernel response of node-centric subgraphs from the sampled graphs in the 3-th line of Fig.\ref{response}. The color of a node indicates the kernel response of its subgraph's similarity to the corresponding graph filter. The color intensity correlates with the strength of the response: yellow indicates a high response, while blue signifies a low response. The graph filters exhibit high consistency with the motifs after model convergence. Besides, it is clear to point out the locations of critical structures within each graph according to the kernel response, offering strong interpretability in predictions.

\subsection{Node Classification}
\label{subsec:Node_Classification}
In this subsection, we evaluate the performance of the proposed model for the node classification task.
\noindent {\bf{Datasets.}} We employ six widely used real-world graph datasets for node classification: Cora, Citeseer, Pubmed\cite{yang2016revisiting}, Chameleon, Squirrel\cite{rozemberczki2021multi}, Actor\cite{pei2020geom}. Each dataset comprises only one graph, and the statistics are summarized in Table.\ref{tab:node_statistics} of Appendix.\ref{app:A}.
\noindent {\bf{Baselines.}} To evaluate the performance of the proposed GOMKCN, we select six state-of-the-art methods as baselines, including GCN\cite{kipf2016semi}, GAT\cite{velivckovic2017graph}, GIN\cite{xu2018powerful}, GraphSAGE\cite{hamilton2017inductive} kerGNN\cite{feng2022kergnns}, and RWK$^{+}$CN\cite{lee2024descriptive}.
\noindent {\bf{Settings.}} Following the default setting in RWK$^{+}$CN\cite{lee2024descriptive}, we randomly split into training, validation, and test sets in a proportion of 6:2:2. Besides, an MLP block is incorporated into the model to perform dimensionality reduction on the raw node features, to eliminate the noise that does not carry meaningful pattern information. GOMKCN is optimized using the Adam optimizer for 200 epochs for all datasets. The learning is set to 0.06 for Cora, CiteSeer, and Actor datasets, while 0.01 for PubMed, Chameleon, and Squirrel datasets. The hyperparameters and their potential values are listed in Table.\ref{tab:node_hyperparameter} of Appendix.\ref{app:A}.

\noindent {\bf{Results.}} We present the predictive accuracy and standard deviations across all evaluated models in Table.\ref{tab:acc_node}, where the best ones are in bold. GOMKCN achieves significantly higher prediction accuracy than other models on the Citeseer, Chameleon, and Squirrel datasets while exhibiting only slight differences in accuracy compared to other models on the Cora, Pubmed, and Actor datasets.

\subsection{Graph Classification}
\label{subsec:Graph_Classification}
In this subsection, we evaluate the performance of the proposed model for the graph classification task.

\noindent {\bf{Datasets.}} We conducted experiments on six widely used datasets, including three biochemical network datasets and three social network datasets: DD\cite{dobson2003distinguishing}, ENZYMES\cite{schomburg2004brenda}, PROTEINS\cite{borgwardt2005protein}, IMDB-B, IMDB-M\cite{yanardag2015deep}, REDDIT-B\cite{yanardag2015deep}. The statistical characteristics of these datasets are shown in Table.\ref{tab:graph_statistics} of Appendix.\ref{app:B}.

\noindent {\bf{Baselines.}} For the graph classification task, we selected seven methods as benchmarks for comparison: GCN\cite{kipf2016semi}, GIN\cite{xu2018powerful}, GraphSAGE\cite{hamilton2017inductive}, TopKPool\cite{gao2019graph}, SAGPool\cite{lee2019self}, kerGNN\cite{feng2022kergnns} and RWK$^{+}$CN\cite{lee2024descriptive}.

\noindent {\bf{Settings.}} For a fair comparison with the baseline methods, we follow the default setting in kerGNN\cite{feng2022kergnns} to adopt the 10-fold cross-validation approach for model evaluation. We share the same dataset index splits, which employ an inner holdout technique with a 9:1 training/validation split for model selection during cross-validation. We use the Adam optimizer to update the model parameters. The model structure is almost the same as that of kerGNN and RWK$^{+}$CN, except for the difference in graph kernels. They utilize an MLP to extract the coupling information from the original node features. Then, the graph kernel convolution is performed to learn decoupled representations for each node. The representation of the entire graph is generated based on the node representations using pooling methods. Another MLP will predict the category of the graph based on this graph representation. The hyperparameters and their potential values are listed in Table.\ref{tab:graph_hyperparameter} of Appendix.\ref{app:B}.

\noindent {\bf{Results.}} We reported the accuracy and standard deviation in Table.\ref{tab:acc_node}, where the best ones are in bold. The prediction results indicate that GOMKCN outperforms seven other models on three datasets. In particular, it achieved significant improvements of 26\% on the ENZYMES dataset.

\begin{table*}[t]
	\caption{Prediction of node classification measured by accuracy (\%).\label{tab:acc_node}}
	\centering
	\setlength{\tabcolsep}{10pt}  
	\fontsize{9}{16}  
	\selectfont
	\begin{tabular}{c|cccccc}
		\hline
		Model & Cora & Citeseer & Pubmed & Chameleon & Squirrel & Actor\\
		\hline
		GCN & {\bf{90.72$\pm$0.3}}(1) & 77.28$\pm$0.5(2) & 88.19$\pm$0.1(4) & 67.06$\pm$1.0(6) & 47.29$\pm$0.6(6) & 30.55$\pm$0.6(4)\\
		\hline
		GAT & 90.02$\pm$0.9(4) & 77.21$\pm$0.5(3) & 87.95$\pm$0.2(7) & 67.94$\pm$1.8(5) & 50.27$\pm$2.5(4) & 29.41$\pm$0.6(5)\\
		\hline
		GIN & 89.13$\pm$0.4(5) & 75.48$\pm$0.2(6) & 89.41$\pm$0.2(3) & 68.68$\pm$0.8(4) & 52.28$\pm$0.9(3) & 28.14$\pm$0.3(7)\\
		\hline
		GraphSAGE & 90.20$\pm$0.5(3) & 77.13$\pm$0.5(4) & {\bf{89.98$\pm$0.1(1)}} & 65.88$\pm$1.3(7) & 44.03$\pm$0.9(7) & {\bf{37.15$\pm$0.6}}(1)\\
		\hline
		kerGNN & 85.80$\pm$0.7(7) & 73.40$\pm$0.5(7) & 88.00$\pm$0.2(6) & 69.70$\pm$0.9(2) & 55.70$\pm$0.4(2) & 28.30$\pm$0.6(6)\\
		\hline
		RWK$^{+}$CN & 88.30$\pm$0.5(6) & 76.70$\pm$0.2(5) & 88.10$\pm$0.2(5) & 69.30$\pm$1.9(3) & 49.70$\pm$1.3(5) & 36.00$\pm$0.2(2)\\
		\hline
		GOMKCN & 90.28$\pm$0.5(2) & {\bf{77.70$\pm$0.5}}(1) & 89.47$\pm$0.2(2) & {\bf{70.24$\pm$1.0}}(1) & {\bf{56.70$\pm$1.0}}(1) & 35.06$\pm$1.1(3)\\
		\hline
	\end{tabular}
\end{table*}

\begin{table*}[t]
	\caption{Prediction of graph classification measured by accuracy (\%).\label{tab:acc_graph}}
	\centering
	\setlength{\tabcolsep}{10pt}  
	\fontsize{9}{16}  
	\selectfont
	\begin{tabular}{c|cccccc}
		\hline
		Models & DD & PROTEINS & ENZYMES & IMDB-B & IMDB-M & REDDIT-B\\
		\hline
		GCN & 72.32$\pm$3.1(4) & 72.69$\pm$3.0(5) & 28.83$\pm$7.4(5) & 72.70$\pm$3.9(4) & 50.33$\pm$4.5(2) & 86.25$\pm$2.1(3)\\
		\hline
		GIN & 72.07$\pm$4.4(5) & 71.87$\pm$4.6(7) & 52.17$\pm$9.6(3) & 72.30$\pm$3.5(6) & 49.80$\pm$4.5(5) & {\bf{90.15$\pm$1.9}}(1)\\
		\hline
		GraphSAGE & {\bf{76.91$\pm$4.6}}(1) & 66.76$\pm$6.4(8) & 20.00$\pm$3.7(8) & 72.90$\pm$4.1(3) & {\bf{51.07$\pm$3.9}}(1) & 62.85$\pm$6.4(7)\\
		\hline
		SAGPool & 70.96$\pm$7.6(8) & 72.23$\pm$3.5(6) & 26.00$\pm$4.5(7) & 72.40$\pm$3.7(5) & 50.13$\pm$3.8(4) & 73.80$\pm$7.2(6)\\
		\hline
		TopK & 71.90$\pm$3.7(6) & 73.22$\pm$5.6(4) & 27.50$\pm$9.7(6) & 72.90$\pm$3.5(2) & 50.27$\pm$4.6(3) & 75.95$\pm$7.9(5)\\
		\hline
		kerGNN & 72.41$\pm$4.4(3) & 74.56$\pm$4.6(2) & 53.00$\pm$7.8(2) & 71.40$\pm$5.2(8) & 47.27$\pm$2.6(7) & 76.20$\pm$3.4(4)\\
		\hline
		RWK$^{+}$CN & 71.56$\pm$3.7(7) & 74.30$\pm$5.3(3) & 50.17$\pm$9.1(4) & 72.10$\pm$4.4(7) & 47.13$\pm$2.7(8) & 59.95$\pm$6.4(8)\\
		\hline
		GOMKCN & 73.25$\pm$3.8(2) & {\bf{75.83$\pm$5.0}}(1) & {\bf{67.00$\pm$6.1}}(1) & {\bf{73.20$\pm$5.6}}(1) & 47.30$\pm$4.0(6) & 88.52$\pm$1.8(2)\\
		\hline
	\end{tabular}
\end{table*}

\subsection{Hyperparameter Analysis}
\label{subsec:Hyperparameter_Analysis}
This subsection analyzes the key parameters that affect model performance, with a particular focus on model depth, the number of subtree levels, and subgraph size.

\begin{figure}[h]
	\centering
	\includegraphics[width=3.5in]{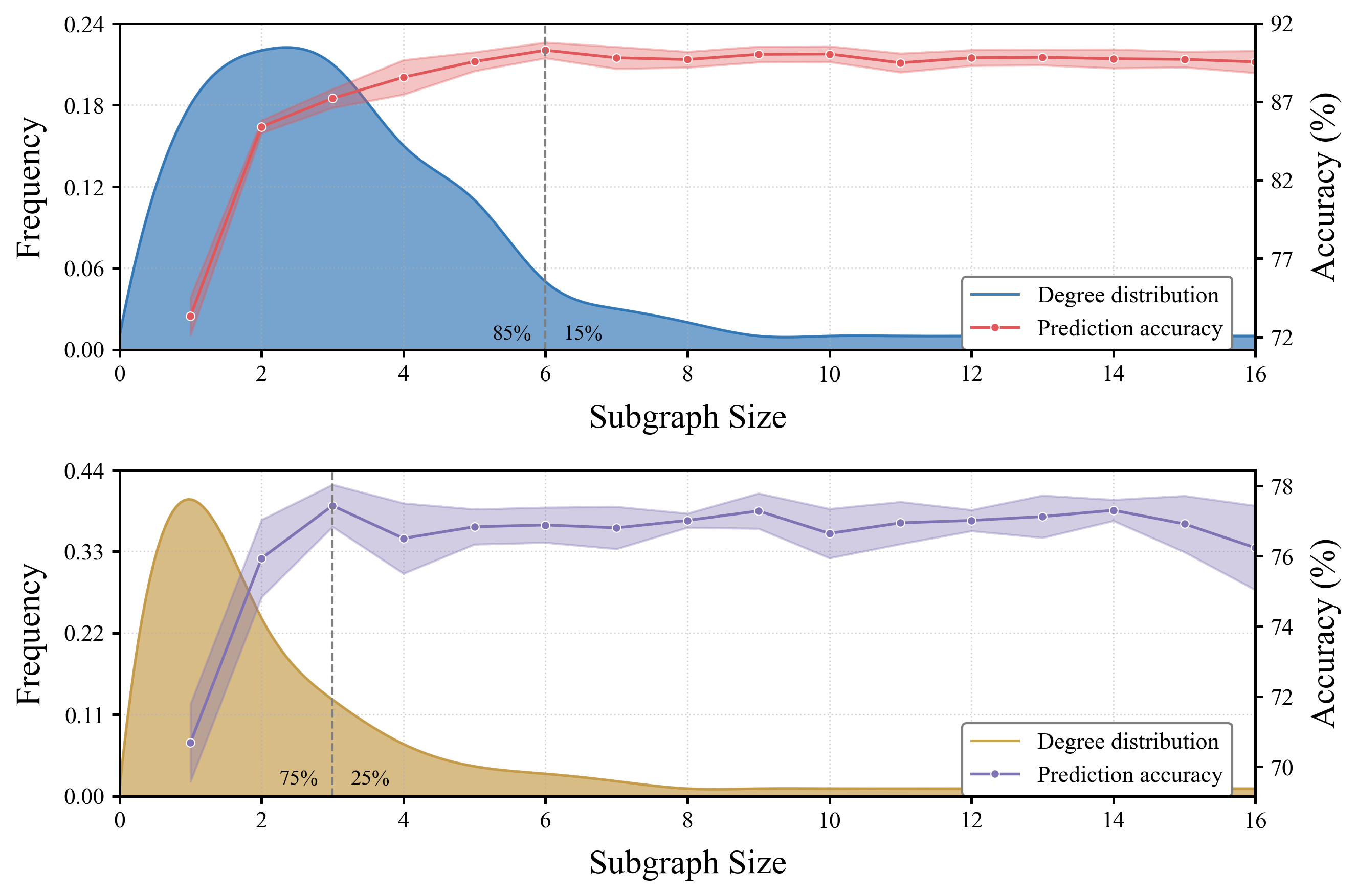}
	\caption{Parameter sensitivity of subgraph size on Cora (top) and Citeseer (button).}
	\label{subgraph_size}
\end{figure}

It is necessary to ascertain the appropriate subgraph size in both subgraph extraction and model configuration. A too small subgraph may not adequately represent the neighborhood information of nodes, while an excessively large subgraph can lead to significant computational complexity. Here, we consider the scenario where predictions are made using the one-hop neighbors of nodes. In Fig.\ref{subgraph_size}, we evaluate the proposed GOMKCN with different sizes of subgraphs from 1 to 16 on Cora (top) and Citeseer (button). Additionally, the degree distribution of nodes in the entire graph is also presented synchronously in the figure. It has been observed that the degrees of nodes in both datasets approximately follow a power-law distribution. The model performance shows a consistent trend on both datasets: the accuracy increases sharply first and then tends to stabilize as the subgraph size grows. When the subgraph size can depict the one-hop subgraphs of 85\% and 75\% of the nodes, the model achieves relatively good prediction performance and has a comparatively low computational complexity.

\begin{figure}[h]
	\centering
	\includegraphics[width=3.5in]{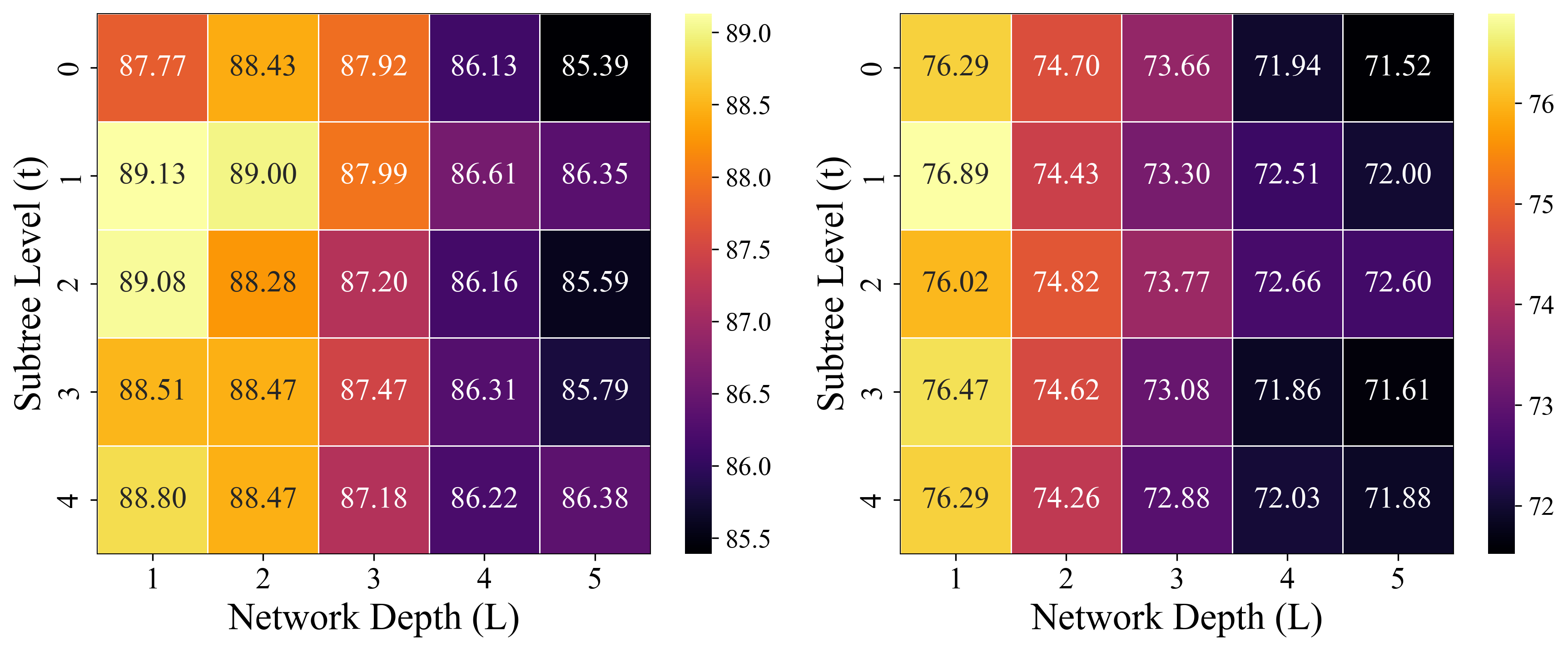}
	\caption{Parameter sensitivity of network depth $L$ and subtree level $t$ on Cora (left) and Citeseer (right).}
	\label{layer_level}
\end{figure}

In Fig.\ref{layer_level}, we evaluate the proposed GOMKCN with different model depth $L$ and subtree level $t$ on Cora (left) and Citeseer (right). Both the model depth varies from 1 to 5, and the subtree level ranges from 0 to 4, and we report the accuracy of prediction under different settings. It has been observed that GOMKCN demonstrates excellent performance even with a relatively small number of network layers and shallow subtree levels. Selecting an appropriate level of subtree to characterize a subgraph can achieve higher accuracy than merely utilizing the original features of the subgraph, possibly because the generated embeddings make fuller use of the topological information of the subgraph. However, as the number of layers in the network and the subtree level continue to increase, the model's performance shows a downward trend, which may be attributed to the inherent overfitting characteristic of the message-passing mechanism. Expanding the network layers and subtree levels is equivalent to enhancing the number of message-passing iterations between subgraphs and within subgraphs. This process resembles a low-pass filtering process, causing the proportion of discriminative information among nodes to become increasingly lower.

\subsection{Time Complexity}
\label{subsec:Time_Complexity}
Assuming that there are $N$ input graphs to be dealt with for disentangled representation. Each input graph contains an average of $n$ nodes with node features of dimension $d$. A $k$ hop subgraph is extracted for each node, where the size is constrained to $m$ nodes by randomly removing peripheral nodes or introducing isolated nodes with zero-feature vectors. The proposed model comprises $L$ convolutional layers, each equipped with $T$ filters that preserve an identical shape to the subgraphs. The subgraphs are encoded for embeddings through $t$ iterations of neighborhood aggregation. The time complexity of the proposed framework can be decomposed into three dominant phases. First, the $k$ hop subgraph extraction for all nodes across $N$ input graphs exhibits a worst-case time complexity of $\mathcal{O}(Nn^{k+1})$. Subsequently, the process of subgraph embedding requires  $\mathcal{O}(NLTntdm^{2})$ operations. Finally, the subtree alignment via optimal matching incurs an $O(NLTnm^{2})$ cost.

\begin{figure}[h]
	\centering
	\includegraphics[width=3.5in]{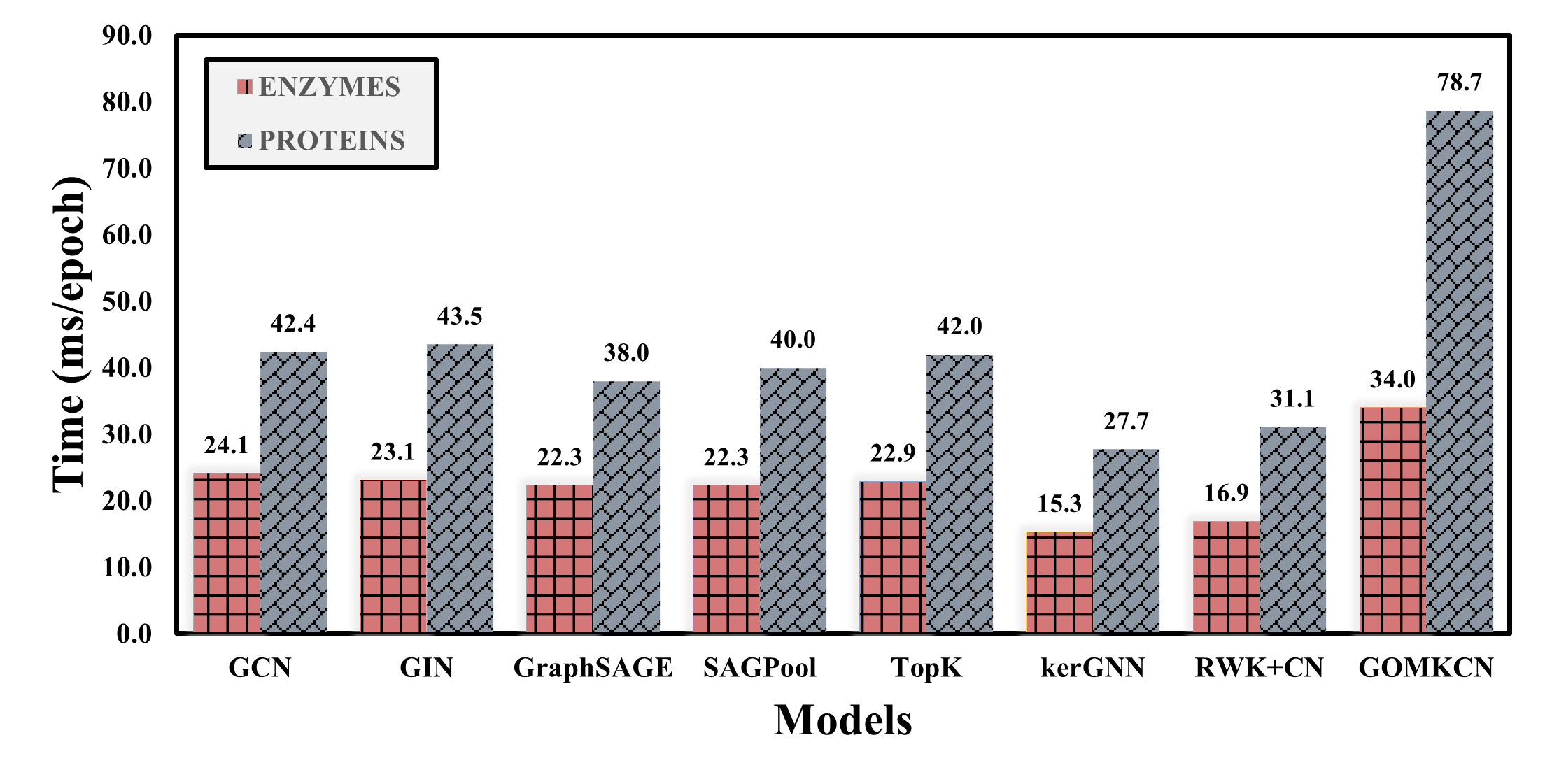}
	\caption{One-epoch running time of models on the ENZYMES and the PROTEINS dataset.}
	\label{time}
\end{figure}

To fully evaluate the runtime of the proposed model, we compare the one-epoch running time with the state-of-the-art methods on the ENZYMES and PROTEINS datasets by averaging the results over 200 epochs, as shown in Fig.\ref{time}. All models were executed on an identical platform with an Intel i9-13900H CPU and NVIDIA GeForce RTX 4090 graphics card. GCN, GraphSAGE, SAGPool, and TopK models have two convolutional layers. For kerGNN, RWK$^{+}$CN, and GOMKCN models, they extract 2-hop node-centric subgraphs and employ a single convolutional layer to generate node representations. These three methods explicitly extract the ego-centered subgraphs, and the time consumption is not included in the picture. Specifically, kerGNN and RWK$^{+}$CN models employ the same subgraph extraction approach, which takes 5.58s and 17.05s on the two benchmark datasets, respectively. In comparison, GOMKCN adopts an optimized implementation that significantly reduces the time of subgraph extraction to 2.68s and 7.17s.

\section{Conclusion}
To address the problem of end-to-end interpretable disentangled representation learning, a novel graph convolutional model, GOMKCN, is designed to learn local graph patterns in graphs automatically. Unlike existing graph convolutional networks, GOMKCN accurately characterizes and identifies the topology and features of node-centric subgraphs with the GOMK and learnable filters. Specifically, a message-passing-based $t$-SE is proposed, which transforms a subgraph or graph filter into a collection of subtrees rooted at individual nodes. The subgraph or graph filter embedding is collectively formed by the subtree embeddings. Additionally, the similarity between a subgraph and a graph filter is measured by a novel OMK, which establishes optimal matching between the subtrees of the subgraph and the filter and evaluates the overall similarity by summing the similarities of subtree pairs. Extensive experimental results demonstrate that the proposed model achieves predictive performance comparable to state-of-the-art methods and exhibits strong interpretability.

Our future work primarily includes the following aspects. On the one hand, GOMKCN mainly focuses on discovering structural patterns in graphs, while the interpretability of node feature patterns remains implicit. Therefore, explicitly interpreting the pattern information in node features is of great significance. On the other hand, the patterns in the graph reflect inherent regularities in graph data. We will strive to leverage this information to defend against adversarial attacks on graph models, thereby achieving robust graph model design.

\newpage

\bibliographystyle{IEEEtran}
\bibliography{references}

\newpage

\begin{appendices}
	
	\section{}
	\label{app:A}
	\begin{table}[h]
		\caption{Dataset statistics for node classification.\label{tab:node_statistics}}
		\centering
		\setlength{\tabcolsep}{5pt}  
		\renewcommand{\arraystretch}{1.5}  
		\begin{tabular}{c|cccccc}
			\hline
			Dataset & Cora & Citeseer & Pubmed & Chameleon & Squirrel & Actor\\
			\hline
			Nodes & 2708 & 3327 & 19717 & 2277 & 5201 & 7600\\
			Edges & 5278 & 4552 & 44324 & 31371 & 198303 & 26659\\
			Features & 1433 & 3703 & 500 & 2325 & 2089 & 932\\
			Classes & 7 & 6 & 3 & 5 & 5 & 5\\
			\hline
		\end{tabular}
	\end{table}

	\begin{table}[h]
			\caption{Range of hyperparameters for node classification.\label{tab:node_hyperparameter}}
			\centering
			\setlength{\tabcolsep}{22pt}  
			\renewcommand{\arraystretch}{1.5}  
			\begin{tabular}{c|c}
					\hline
					output dimension of MLP & [8,16,32,64]\\
					number of filters & [3,5,6,7]\\
					number of nodes in filters & [4,6,8,16,32,80]\\
					hops for subgraph construction & [1,2]\\
					steps for neighbor aggregation & [1,2,3]\\
					width parameter in RBF & 0.6\\
					\hline
				\end{tabular}
		\end{table}
	
	\section{}
	\label{app:B}
	\begin{table}[h]
		\caption{Dataset statistics for graph classification.\label{tab:graph_statistics}}
		\centering
		\setlength{\tabcolsep}{2pt}  
		\renewcommand{\arraystretch}{1.5}  
		\begin{tabular}{c|cccccc}
			\hline
			Dataset & DD & PROTEINS & ENZYMES & IMDB-B & IMDB-M & REDDIT-B\\
			\hline
			Graphs & 1178 & 1113 & 600 & 1000 & 1500 & 2000\\
			Nodes & 334925 & 43471 & 19580 & 19773 & 19502 & 859254\\
			Edges & 843046 & 81044 & 37282 & 96531 & 98903 & 995508\\
			Features & 89 & 32 & 21 & 0 & 0 & 0\\
			Classes & 2 & 2 & 6 & 2 & 3 & 2\\
			\hline
		\end{tabular}
	\end{table}

	\begin{table}[h]
		\caption{Range of hyperparameter for graph classification.\label{tab:graph_hyperparameter}}
		\centering
		\setlength{\tabcolsep}{20pt}  
		\renewcommand{\arraystretch}{1.5}  
		\begin{tabular}{c|c}
			\hline
			epoch & 300\\
			batch size & 128\\
			learning rate & 0.01\\
			number of filters in GOMKCN & [3,4,8,16]\\
			number of node in filters & [4,8,10,16]\\
			dimension of node features in filters & [4,8,16,32]\\
			dropout in MLPs & [0.0,0.2,0.4]\\
			hops for neighborhood construction & [1,2]\\
			steps for neighbor aggregation & [1,2,3]\\
			width parameter in RBFs & 1.0\\
			pooling method & add/mean\\	
			\hline
		\end{tabular}
	\end{table}
\end{appendices}

\end{document}